\documentclass[10pt,twocolumn,letterpaper]{article}
\usepackage{cvpr, times, color, epsfig, graphicx, amsmath, amssymb, enumitem, setspace, authblk, multirow}
\usepackage[pagebackref=false,breaklinks=true,letterpaper=true,colorlinks=true,citecolor=black,linkcolor=black,bookmarks=false]{hyperref}

\cvprfinalcopy

\title{Learning to Point and Count}
\author[1]{Jie Shao}
\author[1]{Dequan Wang}
\author[1]{Xiangyang Xue}
\author[2]{Zheng Zhang}
\affil[1]{Shanghai Key Laboratory of Intelligent Information Processing\\
	
	School of Computer Science, Fudan University}
\affil[2]{Department of Computer Science, New York University Shanghai}
\affil[1]{\tt\small{\{shaojie; dqwang12; xyxue\}@fudan.edu.cn}  \tt\small{$^2$zz@nyu.edu}}

\graphicspath{{img/}}

\begin{document}
\maketitle

\begin{abstract}
	This paper proposes the problem of point-and-count as a test case to break the what-and-where deadlock. Different from the traditional detection problem, the goal is to discover key salient points as a way to localize and count the number of objects simultaneously. We propose two alternatives, one that counts first and then point, and another that works the other way around. Fundamentally, they pivot around whether we solve ``what'' or ``where'' first. We evaluate their performance on dataset that contains multiple instances of the same class, demonstrating the potentials and their synergies. The experiences derive a few important insights that explains why this is a much harder problem than classification, including strong data bias and the inability to deal with object scales robustly in state-of-art convolutional neural networks.
	
	%
	%
\end{abstract}

\section{Introduction}
How do we \emph{point and count}? Seeing an image of a plate of fruits, a fleet of battle ships, a pile of books, a flock of birds, pointing and counting seems effortless, even for children. It is also not a far fetched speculation that such ability could have played an important role in the development of human intelligence. Yet, behind this deceptively simple question lies a profound mystery: what are the necessary building blocks that enable us to pull this off?

Given that deep neural networks share the same fundamental computation unit and the overall architecture as our brain, and that they are proven for their immense classification capability, we set out to extend them and explore designs that can solve the P\&C (Point and Count) problem. In the context of this paper, we restrict ourself to counting and localizing a group of objects of the \emph{same} class distributed in a 2D image.

Even with this restriction, this is a difficult task for today’s deep learning architecture, because it involves solving the ``what'' (classification) and ``where'' (localization) together \cite{kandel2000principles} \cite{ungerleider1994and}. The anatomy of brain suggests that these two functions are embedded in different pathways, but their precise interaction is unknown. Obviously, this deadlock needs to be broken one way or the other.

This paper offers a preliminary study that explores two directions, pivoting around whether we solve ``where'' or ``what'' first. In where-to-what approach, we first learn number of saliency objects from a deep hidden layer (count), and then apply clustering to activation heatmap to localize these objects (point). In what-to-where approach, we record top-ranked features of a given object, and use them as the object's signature to prune heatmap. We then apply clustering to the feature-selected heatmap to simultaneously output both number of objects and their locations. In the context of this paper, we call these two approaches \emph{count-then-point} (C2P) and \emph{point-then-count} (P2C), respectively.

These two approaches are qualitatively different. The P2C proposal is simple, it involves no learning at all, requires only a dictionary built from a pre-trained network, an off-the-shelf clustering algorithm, and can ideally deal with unbounded number of objects. The P2C approach is slightly involved, as it requires additional learning with bounded predictions to come up with counts first. Our experience points out that they are synergistic and can be combined.

If the classification performance has exceeded the human capacity, the seemingly trivial P\&C problem is nowhere close to satisfactory. We perform analysis on the source of failures and identify a number of issues. For instance, we show that the pre-trained network has strong data bias and prefers classes it has seen more often, and that the CNN architecture has trouble dealing with large scale variance. We believe these insights are useful to identify future research directions.

Finally, recognizing that the key challenge for problem like P\&C is the lack of well labelled training data, we propose the
MNIST-LEGO dataset. Essentially, we take individual digits of MNIST and arrange them according to simple construction rules to build new classes of “objects”.
The advantage of such methodology is that we can control the complexity, and can have virtually infinite number of training data to explore the design space quickly. We show how we can replicate the same problems identified in our study of natural images with this dataset.

The rest of the paper is organized as follows. We discuss related work in Section \ref{sec-related}, and then describe the two models in Section \ref{sec-alg}. Experiment results are presented in Section \ref{sec-exp}, MNIST-LEGO is introduced in Section \ref{sec-lego}, and we conclude in Section \ref{sec-conclude}.

\section{Related Work}
\label{sec-related}
Finding the classes of objects and number of instances of each is an important step towards understanding an image. Restricted ourselves to homogeneous objects is one way to divide and conquer the problem space. However, more importantly, this work attempts to explore alternatives that break the what-or-where deadlock.

State-of-art approaches such as \cite{girshickregion} can be adapted to attack this problem, if augmented with additional harnesses to distinguish instances at different location, which may or may not be easy. More importantly, our contribution is to examine the role of appropriate prior to solve the problem without resorting to brute force. The idea of using ``what'' to feature-select heatmap and using saliency count to guide clustering over the heatmap can all be viewed from this perspective.

The proposal of using ``pointers'' to localize objects is potentially a controversial choice. It is stronger than dotting \cite{lempitsky2010learning}, but weaker than bounding box and segmentation. One would argue that it is more natural since salient features are what distinguish an object the most; it would be hard to imagine one needs to draw a bounding box or trace its contour whenever there is a need to point and count. The difficult is in how to evaluate this new way of localization; for the time being we measure how well a pointer is contained within the supplied bounding box to mitigate the issue.

A large body of literature exists on the problem of counting alone. For instance \cite{lempitsky2010learning} treats it as density estimation, and the more recent work from \cite{zhang2014salient} counts by classifying hidden features of convolutional network, which one of our approaches adopts as a building block. Their dataset SOS and MOS provide excellent collection for our study. \cite{segui2015learning} provides an interesting viewpoint of object representation as an indirect learning problem casted as a learning-to-count approach. As we will demonstrate in this paper, solving point and count problem together is much harder. It is also more rewarding, as the experience reveals a number of problems in today's CNN architecture that would otherwise be difficult to discover.

\section{Algorithm}
\label{sec-alg}
\begin{figure*}[htb]
	\begin{center}
		\includegraphics[width=0.8\linewidth]{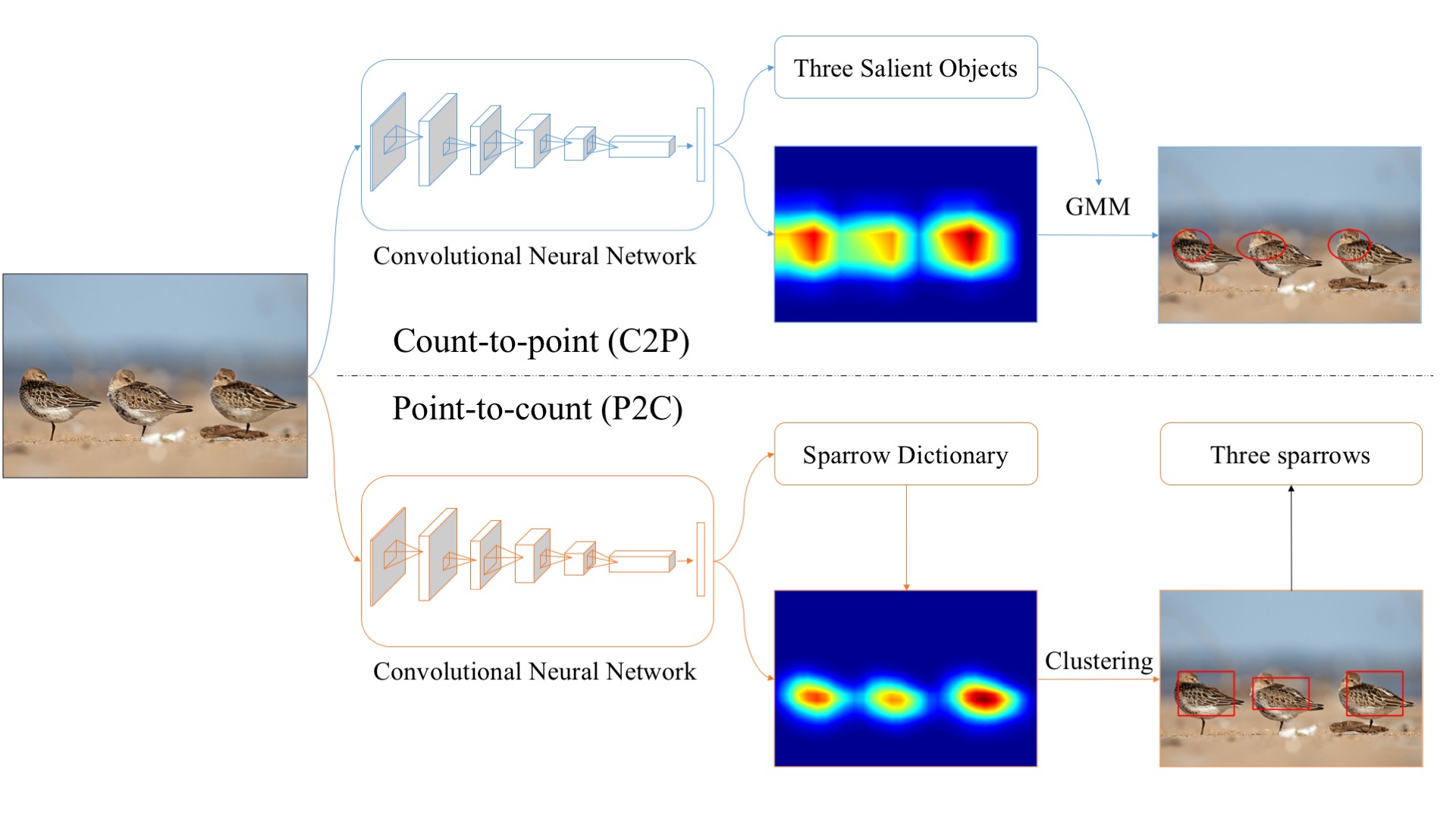}
	\end{center}
	\caption{Overview of our framework. The top is count-then-point (C2P) and the bottom is point-then-count (P2C). Note the pointers of each is different. To output the class also, P2C needs a multi-class classifier, and C2P needs to perform additional classifier after localization. }
	\label{fig:overview}
	\vspace{-3mm}
\end{figure*}

Given an input image $I$, a P\&C algorithm outputs an integer $C$, the number of objects, and a set of \emph{poiners} that locate them. 
The flow of the two algorithms is depicted in Fig\ref{fig:overview} We now proceed to describe each of them in detail.
\label{sec:model}
\subsection{C2P (Count-then-Point)}
The idea of this approach is to first perform localization with the guidance of saliency count. We adopt the proposal in \cite{zhang2014salient}, taking the activation of the last pooling layer and first two fully-connected layers after the convolutional layers in VGG19\cite{simonyan2014very}, and then applying SVM to classify into number of objects. The SVM classifier has bounded output, from zero to $C_{max}$. In the case of the SOS dataset, $C_{max}$ is ``4+'', meaning that $I$ has at least four objects.

It has been well established that moving from input to output layers in a network pre-trained on classification task, activations represent hidden representations that transition from generic to domain-specific\cite{zeiler2014visualizing}. The fully-connected layers collect sophisticated semantic co-occurrence statistic while later convolutional layers gather abounding spacial information. To handle layers with different scales of magnitude, each layer output is normalized before concatenating feature space.

The output of SVM, the predicted counting numbers $C$, feeds into a GMM (Gaussian Mixture Model), along with a properly normalized heatmap generated from the last convolutional layer. Here GMM is a collection of $C$ Gaussian distribution and each distribution represents a cluster of data points. The pointer in this approach is generated by plotting ellipses corresponding to the equation $\{ \mathbf{x}: (\mathbf{x}-\mu_c)^\top \Sigma_c^{-1} (\mathbf{x}-\mu_c) = 1 \}$ where $\mu_c$ denotes the mean, $\sigma_c$ represents diagonal covariance matrice and $\pi_c$ is the prior probability for each of the modes.


\subsection{P2C (Point-then-Count)}

This pipeline is rooted at the belief that to point and count multiple instances we should start with understanding their class first. During the inference stage of a DNN-based classifier, the ``signature'' of the class can prune away distracting activations, leaving only those of the objects we are looking for.

We implement the signature as a dictionary, indexed by object class. The entry $dict[K]$ stores the ranking of feature maps, in the form of their layers and indices of a pre-trained classifier when it infers objects of class $K$. The ranking score is the activation levels of the feature maps when seeing objects of class $K$; higher indicates more relevance to the class.

Upon receiving $I$, the output of the classifier identifies the classes embedded in $I$. In the context of this work, the output identifies only one class. In the extended scenario where $I$ contains objects of multiple different classes, a classifier that is capable of output multiple objects \cite{wei2014cnn}\cite{gong2013deep} can be used, leading to a nested iterative process: the outer loop goes through the classes, whereas the inner loop look for instances of the class.

The algorithm now uses $K$ to retrieve the top ranked features from the dictionary. A \emph{feature-selected} heatmap is produced, by stacking together activations of the selected feature maps. After normalization steps similar to the count-then-point approach, we apply an off-the-shelf clustering algorithm, the number of clusters and the points belong to the clusters are output simultaneously. In P2C, a pointer is the smallest bounding box that contain all points of a cluster.

%
%

\section{Experiment}
\label{sec-exp}
This section presents our preliminary results into the P\&C problem. We will begin with overall performance of the two algorithms, followed by detailed analysis of where they fall short. Our results indicate challenges in both data distribution as well as inherent problems in CNN.

\textbf{Dataset} The SOS\cite{zhang2014salient} dataset contains 6,900 images labeled as 0, 1, 2, 3 or 4+ salient objects. For fair comparison, we use standard dataset split and report results in terms of classification accuracy. The training set consists of 5,520 images and test set consists of 1,380 images. The MSO dataset contains 1,224 images out of 1,380 images from the test set of the SOS datasets. Each image in MSO has bounding box annotations for each individual salient object, and most of the time they are of the same instances, makes it suitable to evaluate P\&C. These two datasets have more balanced proportions of samples from zero to multiple salient objects, simulating realistic challenging scenarios.

\textbf{Network configuration}.

\textbf{P2C} We use the publicly available trained VGG16\cite{simonyan2014very} CNN model.
Given an image from category $k$, two 512-d vectors were obtained by implementing max pooling over the 512 feature maps of the last two convolutional layer, they are sorted according to average activations and inserted into the dictionary. 
During the inference stage, the output label is used to consult the dictionary. We select the top 20 and 50 features from the last and its previous layers, their respective activations are stacked together. This feature-selected heatmap is then normalized by subtracting 2x of the global mean. We use the clustering algorithm from \cite{rodriguez2014clustering} to output both count and the pointers. 

\textbf{C2P} We adapt the publicly available VGG19\cite{simonyan2014very} model, finetuned 20,000 classes ImageNet\cite{deng2009imagenet} images with around 1,000 samples per class. At pre-training phase we directly replace the 1,000 output softmax layer with 20,000 neurons and train around 430,000 iterations with 128 mini-batches. We choose this more powerful network in part to understand what more gains it can have over the approach in \cite{zhang2014salient}. We do not believe it will qualitatively change the results but will experiment with the same network as used in P2C in the future. Other steps are already described in Section \ref{sec:model}.

\subsection{Overall results}

\begin{table}
	\begin{center}
		\begin{tabular}{c|c c c c c|c}
			\hline
			\hline
			Methods & 0 & 1 & 2 & 3 & 4+ & mean(\%) \\
			\hline
			Chance \cite{zhang2014salient} & 28 & 48 & 19 & 12 & 7 & 23\\
			SalCount \cite{zhang2014salient} & - & 55 & 21 & 16 & 11 & - \\
			Zhang \etal \cite{zhang2014salient} & 93 & 90 & 51 & 48 & 65 & 69\\
			\hline
			C2P & 89 & 83 & \bf59 & \bf50 & \bf72 & \bf71\\
			P2C & 52 & 75 & 35 & 42 & 44 & 50\\
			\hline
			\hline
		\end{tabular}
		\vspace{1mm}
		\caption{Quantitative results on the SOS dataset \cite{zhang2014salient} in comparison with state-of-the-art methods. }
		\label{tab:SOS}
	\end{center}
	
	\begin{center}
		\begin{tabular}{c|c|c c c}
			\hline
			\hline
			Overlap&Methods & 1 & 2 & 3 \\
			\hline
			\multirow{2}{*}{0.5} &
			C2P & .87 & .86 & .84 \\
			& P2C & \bf.92 & \bf.97 & \bf.95\\
			\hline
			\multirow{2}{*}{0.8} &
			C2P & .76 & .70 & .63 \\
			& P2C & \bf.81 & \bf.86 & \bf.85\\
			\hline
			\hline
		\end{tabular}
	\end{center}
	\vspace{1mm}
	\caption{C2P and P2C pointing accuracy, as a function of pointer overlap ratio. 1, 2 and 3 stand for SOS categories.}
	\label{tab:pointing}
	%
\end{table}


Table-\ref{tab:Success} presents some of the examples where both count and pointers are correct. The overall counting performance on SOS is reported in Table-\ref{tab:SOS}. As C2P takes essentially the same strategy as \cite{zhang2014salient}, its performance is also similar. P2C is designed to detect objects one class at a time and SOS contains heterogeneous classes, its low accuracy is expected.

\begin{figure}
	\centering
	\label{FIXME}
	\begin{minipage}{0.5\textwidth}
		\includegraphics[width=.8\textwidth]{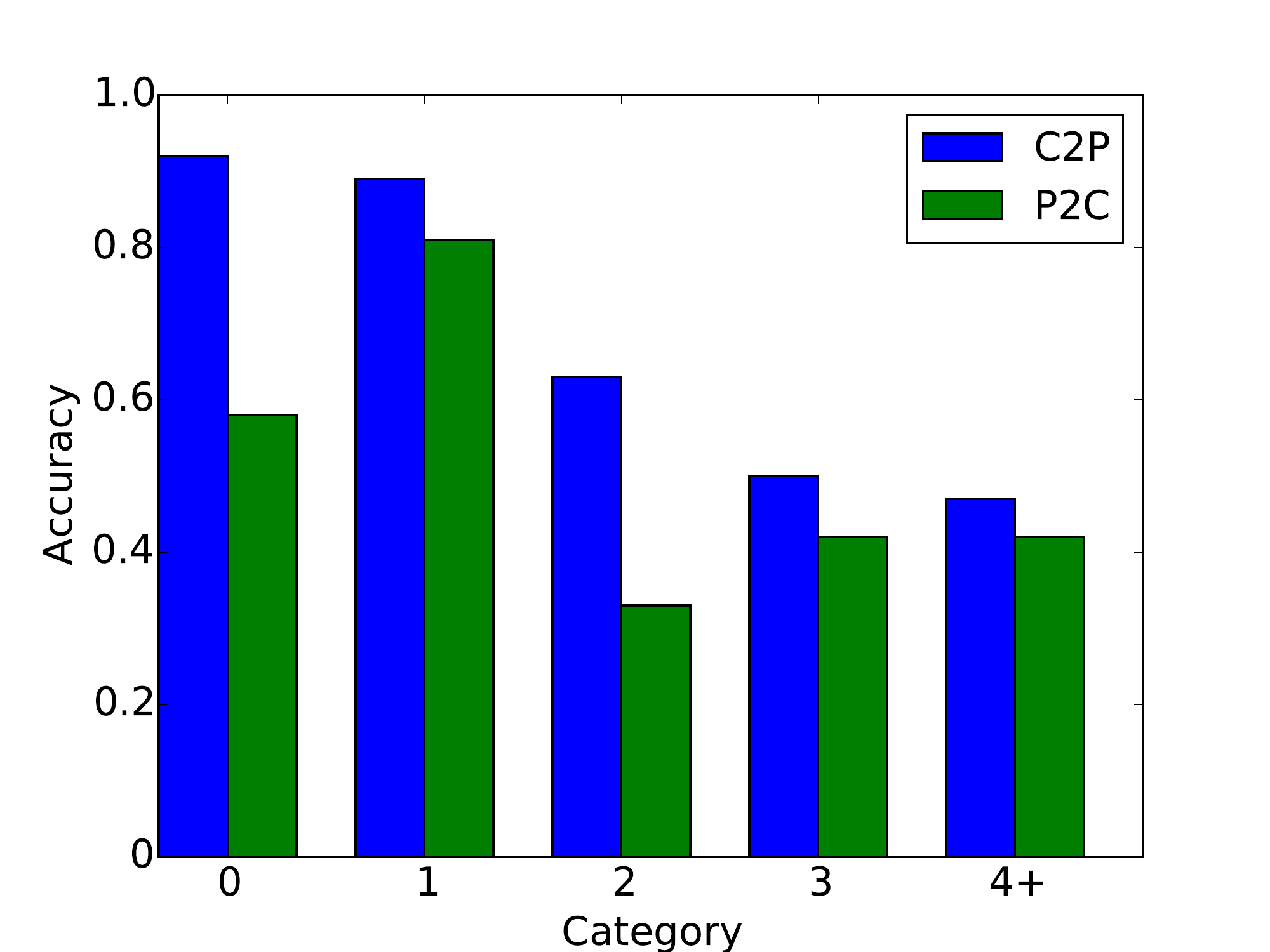}
		\centering
		\caption{C2P versus P2C on count accuracy in MSO}
		\label{fig-count}
	\end{minipage}
	
	\begin{minipage}{0.5\textwidth}
		\includegraphics[width=.8\textwidth]{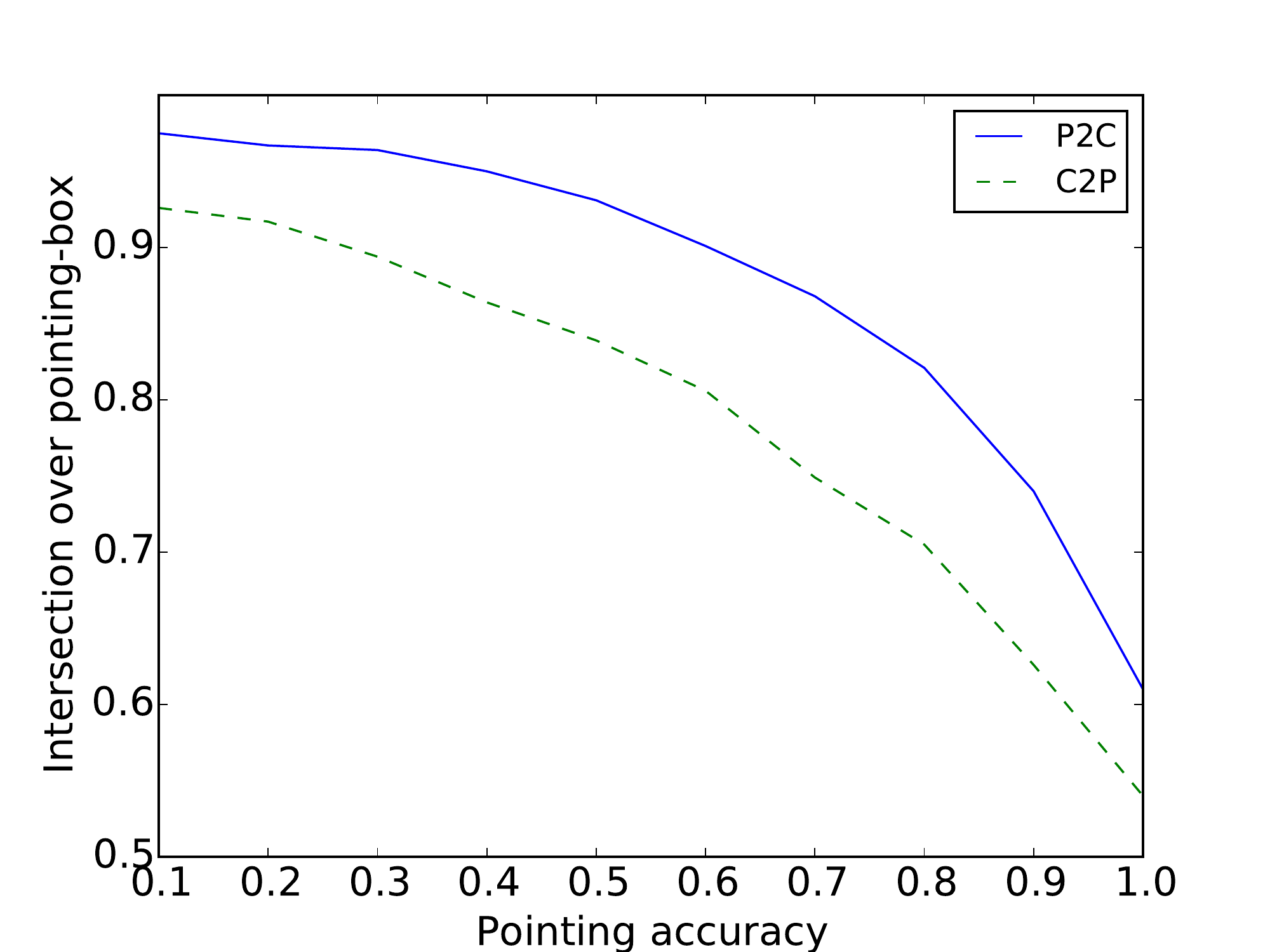}
		\centering
		\caption{Pointing accuracy compared, as a function of pointer overlap ratio}
		\label{fig-point}
	\end{minipage}

\end{figure}

It is more meaningful to compare the two methods on the MSO dataset, as shown in Figure \ref{fig-count}. P2C has a naive treatment for zero objects, so for that category it does not perform well. Otherwise, P2C is close to C2P. However, the performance on pointing is reversed.

\begin{table*}[ht]
	\small
	\begin{center}
		\begin{tabular}{ccccc}
			Ground Truth & C2P Heatmap & C2P Output &  P2C Heatmap & P2C Output \\
			\includegraphics[width=0.10\linewidth]{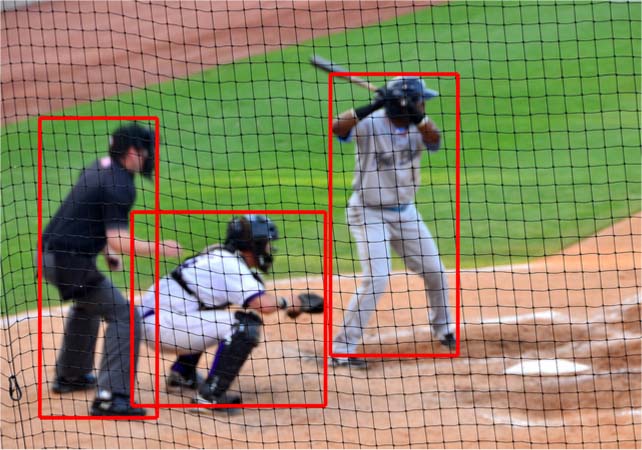}&
			\includegraphics[width=0.10\linewidth]{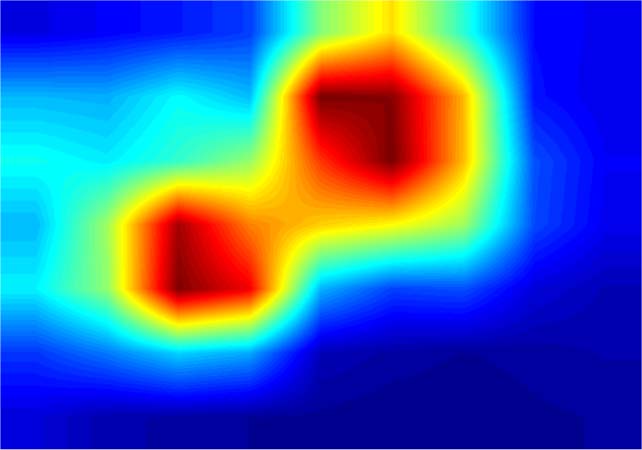}&
			\includegraphics[width=0.10\linewidth]{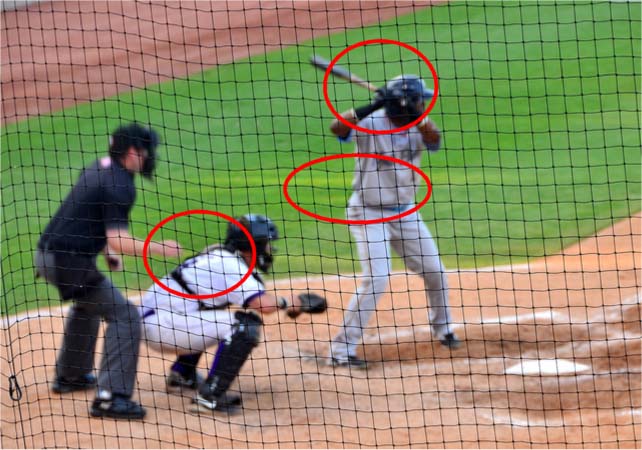}&
			\includegraphics[width=0.10\linewidth]{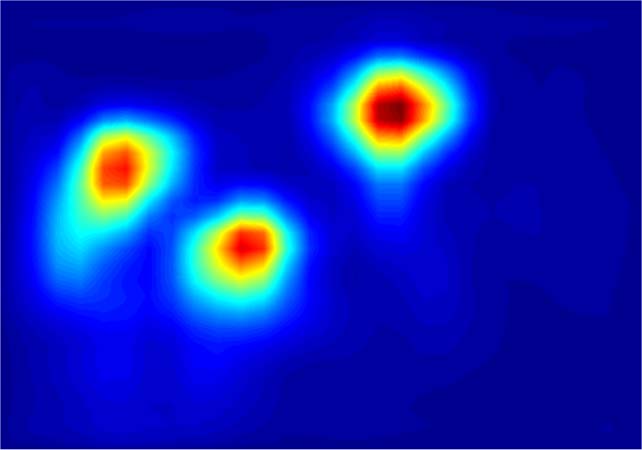}&
			\includegraphics[width=0.10\linewidth]{COCO_COCO_train2014_000000079286_5}\\
			
			\includegraphics[width=0.10\linewidth]{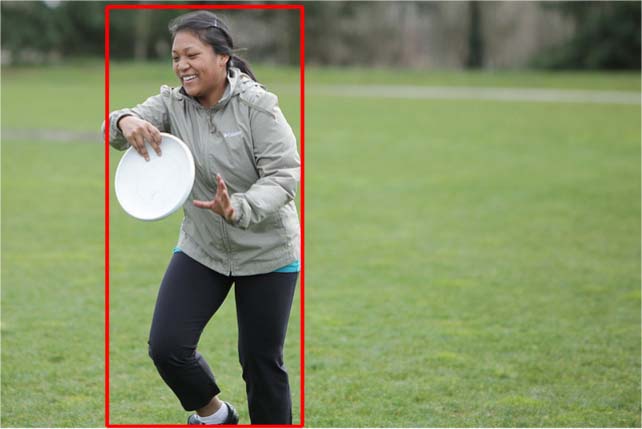}&
			\includegraphics[width=0.10\linewidth]{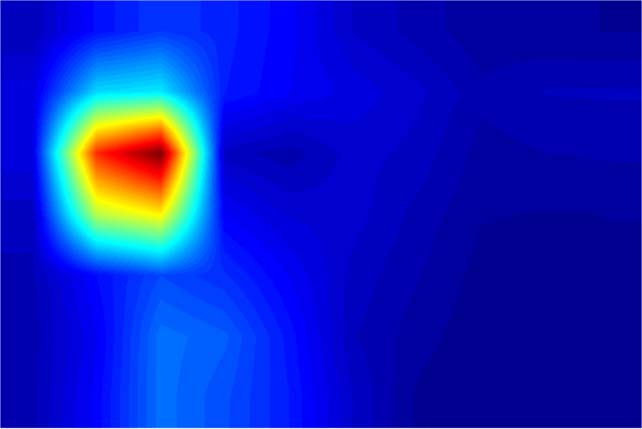}&
			\includegraphics[width=0.10\linewidth]{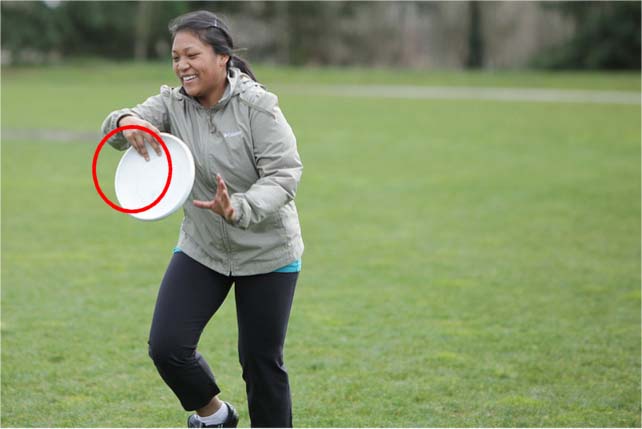}&
			\includegraphics[width=0.10\linewidth]{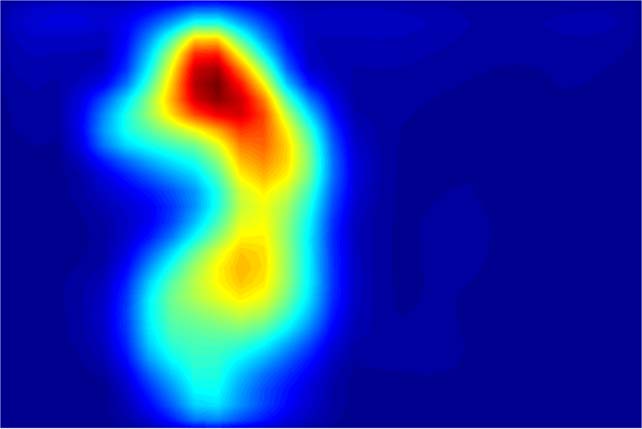}&
			\includegraphics[width=0.10\linewidth]{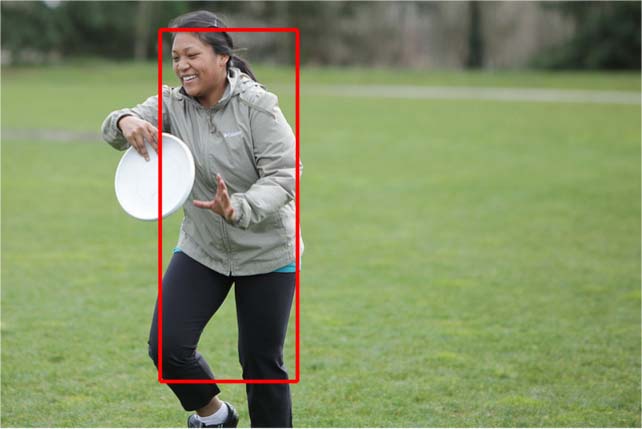}\\

		\end{tabular}
	\end{center}
	\vspace{-3mm}
	\caption{Feature-selected heatmap in P2C out-performs C2P in localization}
	\label{tab:p2c-c2p-pointing}
\end{table*}

To evaluate pointing accuracy, we iterate through the ground-truth bounding boxes in MSO images, each time taking away a pointer that has more than a specified overlap ratio. Figure \ref{fig-point} reports the accuracies of all cases where the total count is correct. P2C out-performs C2P in all cases, especially significant for larger counts and when the overlap ratio is higher (meaning more accurate localization-wise). Table \ref{tab:pointing} gives the breakdown according to number of objects. Upon close inspections, we see that P2C, which operates on feature-selected heatmap, has a better chance to localize correctly (Table \ref{tab:p2c-c2p-pointing}).




\subsection{Failure analysis}

There are roughly 10\% cases (around 150 images) where both algorithms fail. We manually inspect all of them and categorize the errors. The breakdown is included in the supplementary materials, and a subset is used to illustrate the problems. 

\textbf{Data bias}. There is a strong tendency that the network prefers classes of object that it has seen more often. As our network is trained on ImageNet data, generic human bodies is not what it is expecting, and is often ignored. This effect is more pronounced when human figures do not dominate the image. Cases shown in Table \ref{tab:Data} support this conclusion.
In the fourth image, the slight color distortion at the corners leads the network to bias towards ``TV'', ignoring the people even when they are in the center of the image.

The rest of the errors seem to scatter everywhere. However, there is a common thread: the inability of CNN to deal with object scales robustly. There have been moderate steps towards solving this problem ~\cite{xu2014scale, kanazawa2014locally}, but the magnitude of the issue seems to exceed their capabilities.

\textbf{Big-O}. Big objects causes both over- and under-count. In these cases, the objects are big enough that there are no meaningful distinction between foreground and background. When the features are sharp and local, our clustering algorithm is confused and it assigns high-lighted areas of the heatmap into different clusters, resulting over-count (Table \ref{tab:BigO}). On the other hand, if the features are more distributed, normalization reduces the heat uniformly, resulting in under-count (not shown). Unfortunately, this happens even though we classify the object correctly. However, this is a case where C2P does well: the clustering algorithm has an easy time to deal with one object than many.

\textbf{Parallel suppression}. When objects of moderate sizes are close, but not yet close enough to group their heatmaps, there is a peculiar fact that only one of them has a complete heatmap; others are disrupted one way or the other (Table \ref{tab:Parallel}). This maybe related to the rigid structure of receptive fields (RF) in CNN. RFs are square, and real objects are seldom so regular. We hypothesize that, during training, kernels with large RFs have achieved partial robustness when multiple objects are included in their views. The net effect is that it has spatial preferences. This asymmetric property is fine for classification, but is damaging for localization through heatmap saliency. As it so happens, this effect can cause both under- and over-count.

\textbf{Small objects and close-by confusion}. This issue does not come as a surprise: when objects are small, the kernels with matching RFs are in lower layers where the capacity to capture complexity is still low. When they propagate towards the higher layers, the RFs are already too big. The result is that the heatmap of individual objects are tangled together. Consequently, small and close-by confusion almost inevitably leads to under-count (Table \ref{tab:Closeby}).

\subsection{Which goes first: point, or count?}

The point-then-count proposal (P2C) is simple: there is no additional learning required. As a side benefit, it can point and count unbounded number of objects. It is also elegant in the sense that points and count are output simultaneously. Its weakness is placing too much trust on the clustering algorithm.

In contrast, count-then-point (C2P) is slightly more involved as it requires training to count, and only count bounded objects. Counting first makes the clustering more informative, this is why C2P wins against P2C when one object dominates. However, without a prior of what object one is searching for, the heatmap mix together saliency of all objects that the network thinks to be interesting, depriving clustering from another source of information. It is therefore interesting to see how these two approaches can be combined.

%
%
%

\section{MNIST-LEGO}
\label{sec-lego}
We have shown that tackling problems such as P\&C is a much more difficult than simple classification. In contrast to the growing optimism that most if not all problems can be solved with today's large collection of weakly labeled real-world images, our investigation suggests otherwise. In contrast to well above ten-million ImageNet data\cite{russakovsky2014imagenet}, SOS has merely 5,000, and our experience is that the label ambiguity can be an issue. The ideal dataset is where we can label not only the class but also their many instances and, as we have argued earlier, the change of scales. The combination is daunting.

Dataset of the reduced complexity, such as MNIST, seems to have fallen out of favor, as they are not \emph{real enough}. They can be improvised so we have a mid-ground to attack more complex problems. This approach has been used to understand how to build robust noise-free features (e.g. MNIST-back-img), or decode how top-down attention/belief plays a role in disentangle highly occluded objects (e.g. MNIST-2\cite{wang2014attentional}), or attention through sequencing (e.g. DeepMind\cite{mnih2014recurrent}).

Partially inspired by the Deformable Parts Model and the problems we discovered in this study, we synthesize the MNIST-LEGO dataset. The idea is to treat each digit as a part, and stack them together according to some rules. The stacked digits can form classes of “objects” of arbitrary complexity. The number of digits forming the new class, the shape, the orientation and, if needed, lighting, can all be controlled. Furthermore, we can add structured background noises as well. The number of training data is big enough to tune new algorithms: if there are $N$ MNIST digit samples, and a new class is made up by $K$ digits, then there is a total of $O(N^K)$ training data for this class.

Table \ref{tab:mnist-lego} shows some samples of the MNIST-LEGO. Running the P2C algorithm generates  qualitatively identical problems as we have seen in the real-world images (Table \ref{tab:failure-mnist-lego}). This dataset has 100 3-digit classes with image-background noise. A 4-layer convolutional neural network, plus one NIN (network in network)\cite{lin2013network}, and a global average pooling layer. The convolutional layers each contain 32 to 256 kernels of size $5\times5$. The network tested has already reached $85.4\%$ top-1 accuracy (state-of-art CNN on ImageNet 1K is around $72\%$). 

As they are easy to generate and can be learned on a smaller network, they can be used to investigate new and transferable solutions.

\begin{table*}[ht]
	\small
	\begin{center}
		\begin{tabular}{ccccc}
			\includegraphics[width=0.10\linewidth]{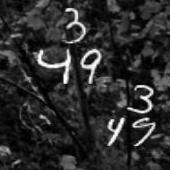}&
			\includegraphics[width=0.10\linewidth]{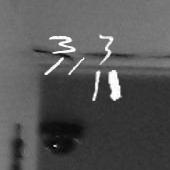}&
			\includegraphics[width=0.10\linewidth]{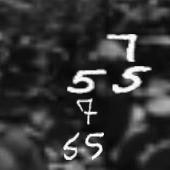}&
			\includegraphics[width=0.10\linewidth]{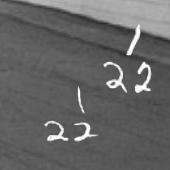}&
			\includegraphics[width=0.10\linewidth]{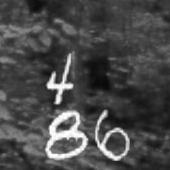}\\

		\end{tabular}
	\end{center}
	\vspace{-3mm}
	\caption{Samples of MNIST-LEGO with image-background noise}
	\label{tab:mnist-lego}
\end{table*}
\begin{table*}[ht]
	\small
	\begin{center}
		\begin{tabular}{ccccc}
			Ground Truth &  P2C Heatmap & P2C Output \\
			\includegraphics[width=0.10\linewidth]{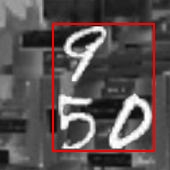}&
			\includegraphics[width=0.10\linewidth]{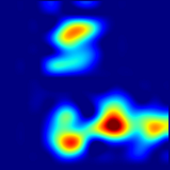}&
			\includegraphics[width=0.10\linewidth]{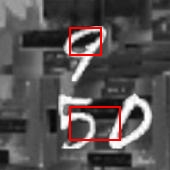}&\\
			
			\includegraphics[width=0.10\linewidth]{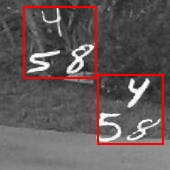}&
			\includegraphics[width=0.10\linewidth]{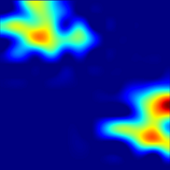}&
			\includegraphics[width=0.10\linewidth]{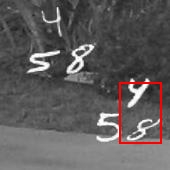}&\\
			
			\includegraphics[width=0.10\linewidth]{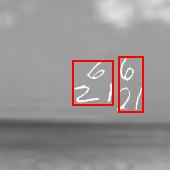}&
			\includegraphics[width=0.10\linewidth]{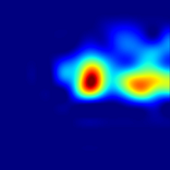}&
			\includegraphics[width=0.10\linewidth]{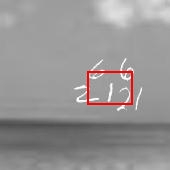}&\\
			
		\end{tabular}
	\end{center}
	\vspace{-3mm}
	\caption{Big-O, parallel suppression and close-by failures replicated in 3-digit MNIST-LEGO}
	\label{tab:failure-mnist-lego}
\end{table*}
%
%
%

\section{Conclusion}
\label{sec-conclude}
Solving the what-and-where problem is a difficult challenge. We believe it is time to take a more holistic perspective, ranging from fabricating new dataset, to employing intuitive and important prior. The work presented in this paper is only a modest step. 

In terms of strategy, it is wise to divide and conquer the problem space. Building a robust per-class detector, combining it with bottom-up proposals and then sliding a bounding box is a popular approach in this direction. We follow the same spirit, but suggest a different tactic. After thoroughly understanding how to combine point-then-count and count-then-point, a promising next step is to take it as a building bock and integrate with a multi-label CNN, making a totally integrated system.

\begin{spacing}{0.95}
{\small
	\bibliographystyle{ieee}
	\bibliography{arXiv}
}
\end{spacing}

\begin{table*}[ht]
	\small
	\begin{center}
		\begin{tabular}{ccccc}
			Ground Truth & C2P Heatmap & C2P Output &  P2C Heatmap & P2C Output \\

			\includegraphics[width=0.10\linewidth]{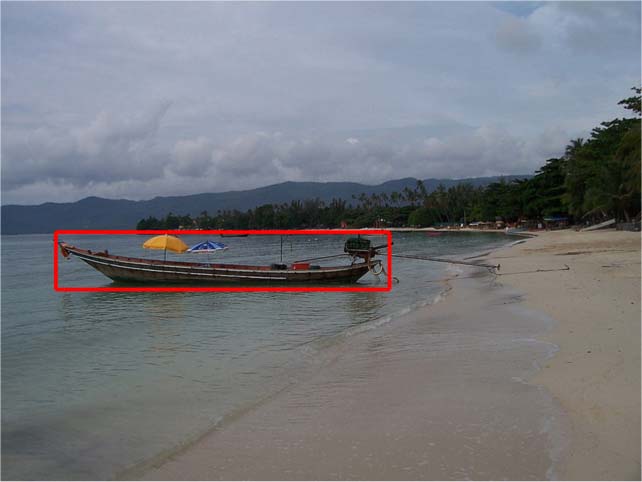}&
			\includegraphics[width=0.10\linewidth]{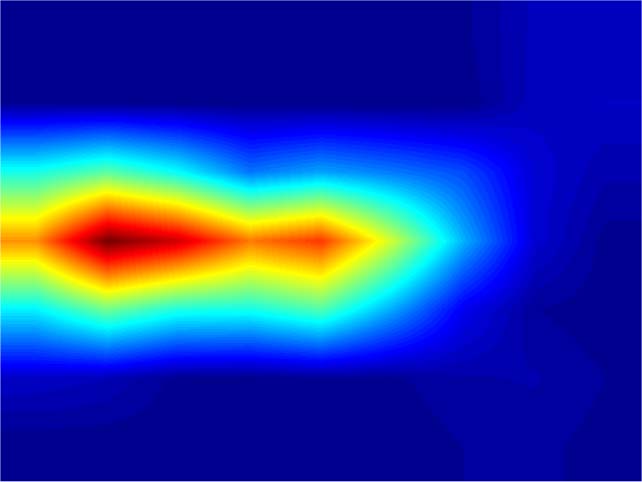}&
			\includegraphics[width=0.10\linewidth]{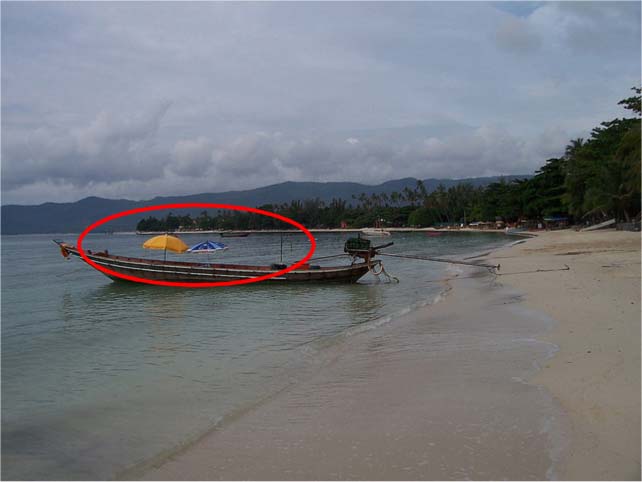}&
			\includegraphics[width=0.10\linewidth]{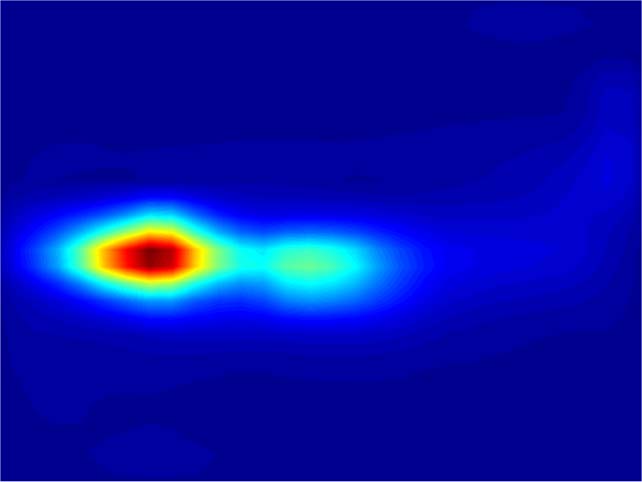}&
			\includegraphics[width=0.10\linewidth]{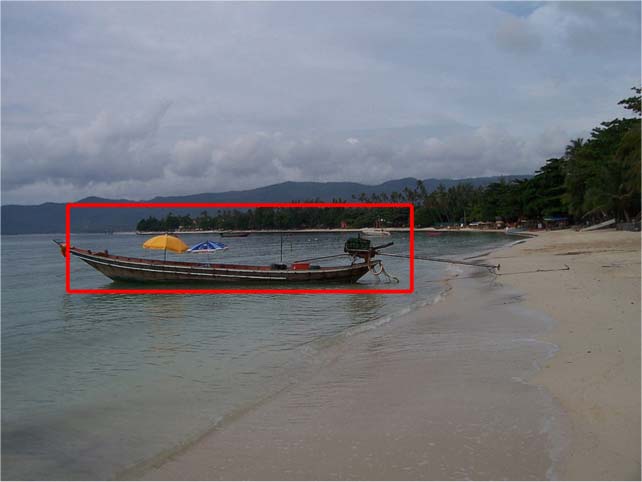}\\
			
			\includegraphics[width=0.10\linewidth]{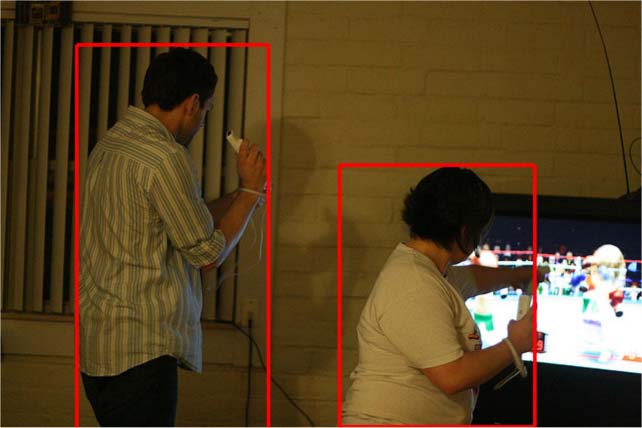}&
			\includegraphics[width=0.10\linewidth]{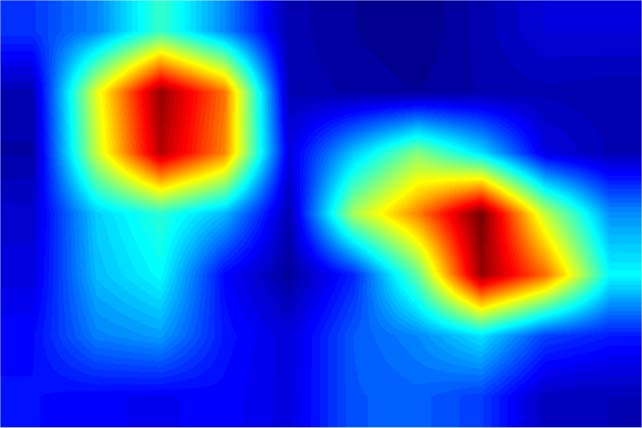}&
			\includegraphics[width=0.10\linewidth]{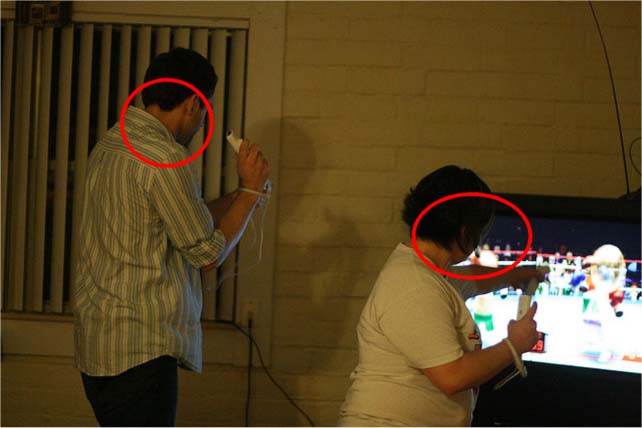}&
			\includegraphics[width=0.10\linewidth]{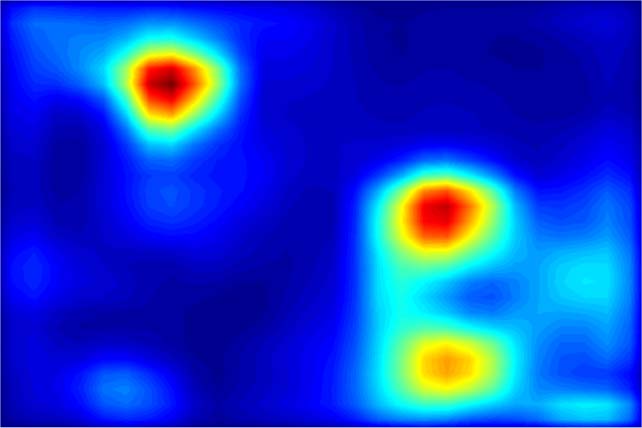}&
			\includegraphics[width=0.10\linewidth]{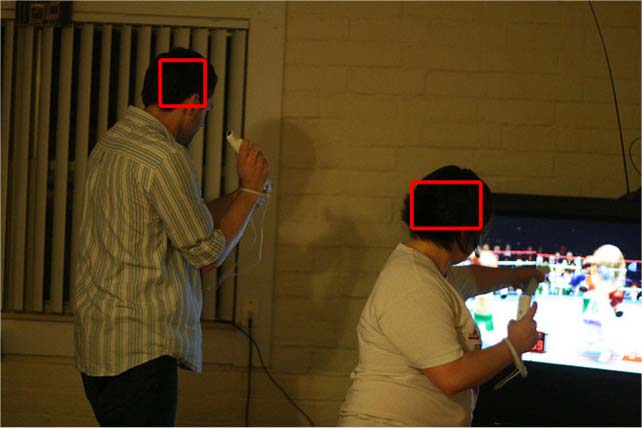}\\
			
			\includegraphics[width=0.10\linewidth]{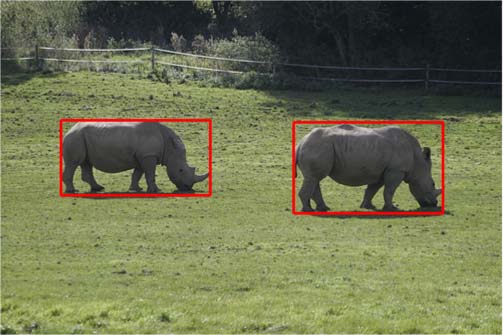}&
			\includegraphics[width=0.10\linewidth]{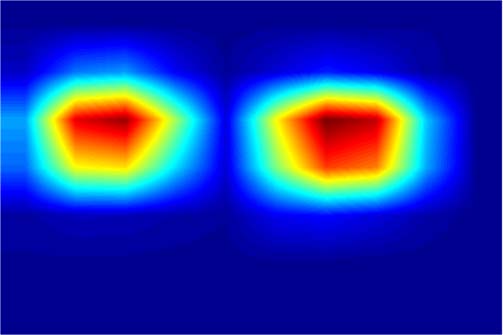}&
			\includegraphics[width=0.10\linewidth]{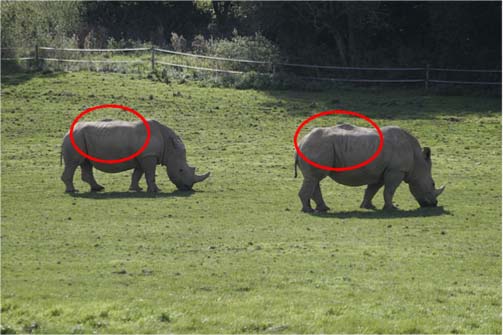}&
			\includegraphics[width=0.10\linewidth]{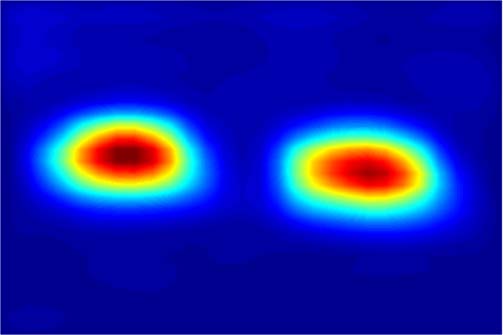}&
			\includegraphics[width=0.10\linewidth]{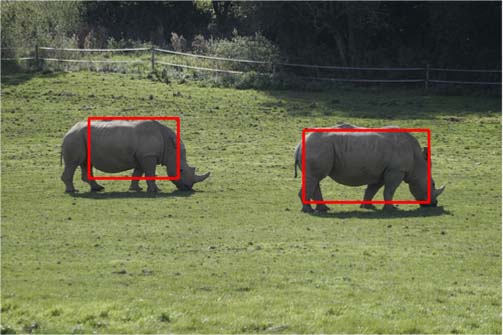}\\

			\includegraphics[width=0.10\linewidth]{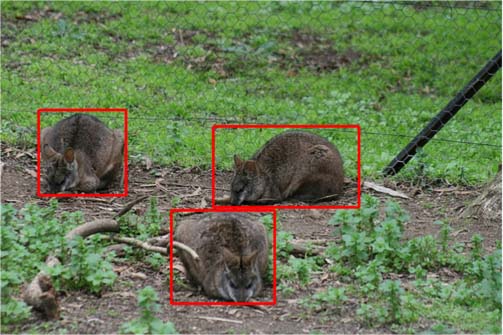}&
			\includegraphics[width=0.10\linewidth]{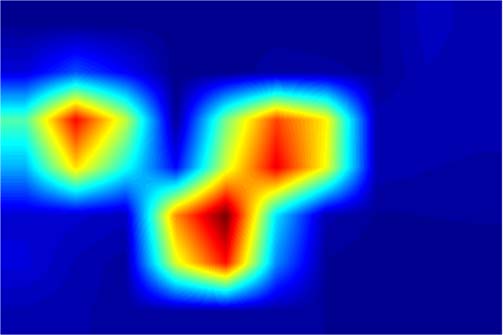}&
			\includegraphics[width=0.10\linewidth]{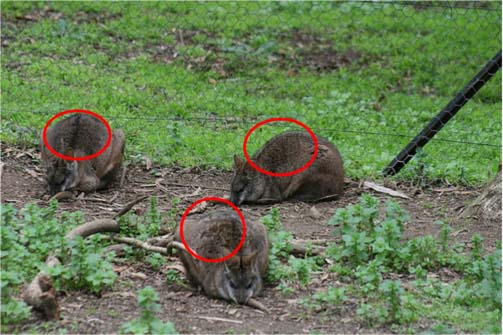}&
			\includegraphics[width=0.10\linewidth]{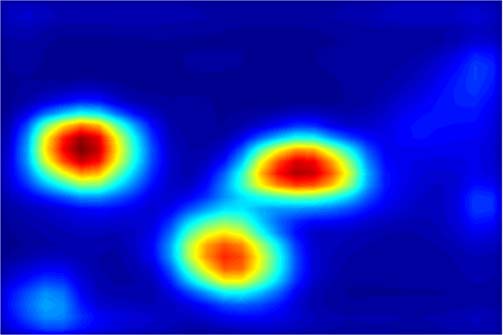}&
			\includegraphics[width=0.10\linewidth]{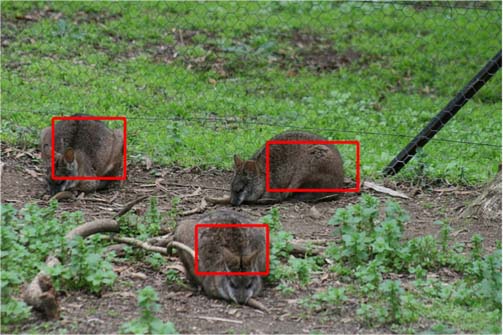}\\
			
			\includegraphics[width=0.10\linewidth]{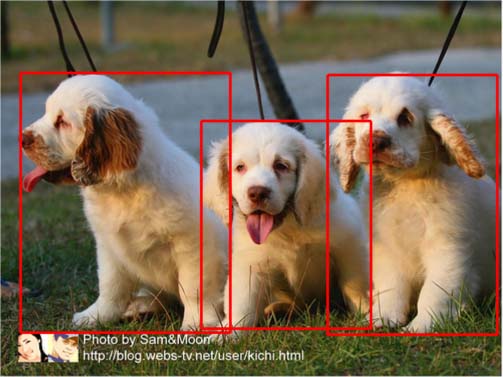}&
			\includegraphics[width=0.10\linewidth]{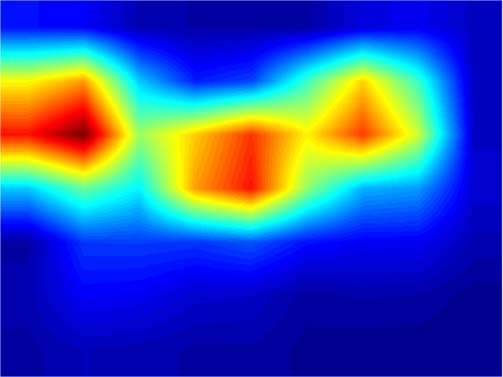}&
			\includegraphics[width=0.10\linewidth]{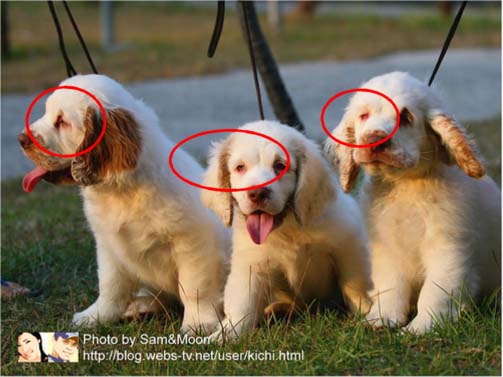}&
			\includegraphics[width=0.10\linewidth]{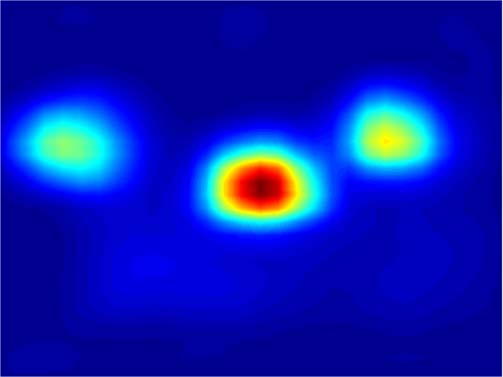}&
			\includegraphics[width=0.10\linewidth]{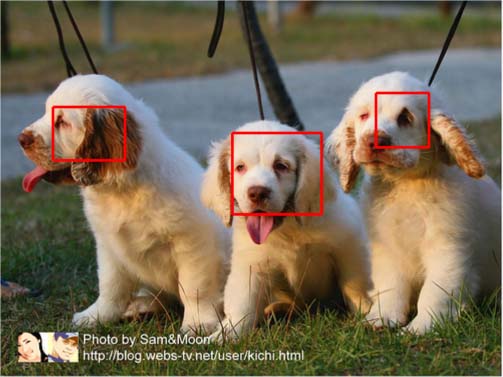}\\

		\end{tabular}
	\end{center}
	\vspace{-3mm}
	\caption{Examples on MSO dataset\cite{zhang2014salient} where both P2C and P2C are successful}
	\label{tab:Success}
\end{table*}

\begin{table*}[ht]
	\small
	\begin{center}
		\begin{tabular}{ccccc}
			Ground Truth & C2P Heatmap & C2P Output &  P2C Heatmap & P2C Output \\
			\includegraphics[width=0.10\linewidth]{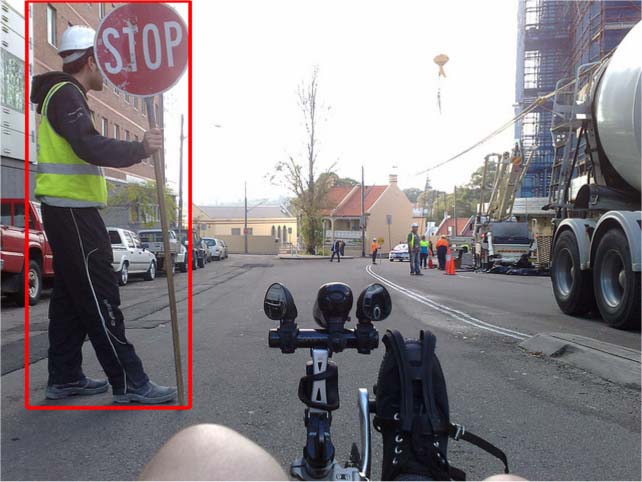}&
			\includegraphics[width=0.10\linewidth]{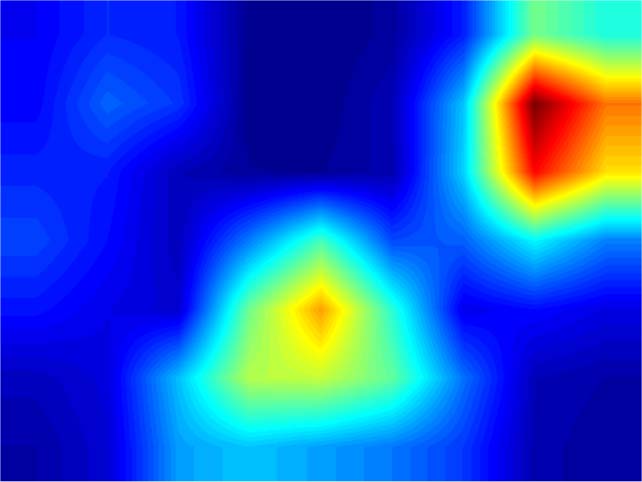}&
			\includegraphics[width=0.10\linewidth]{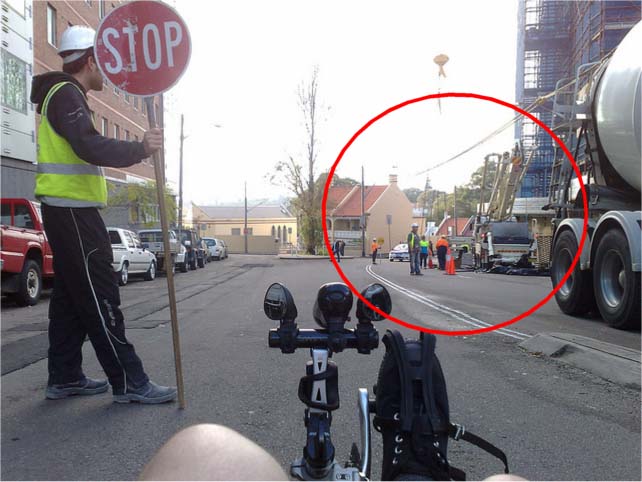}&
			\includegraphics[width=0.10\linewidth]{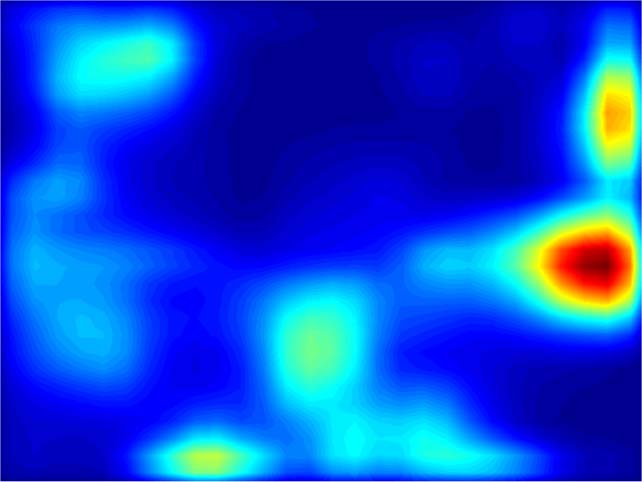}&
			\includegraphics[width=0.10\linewidth]{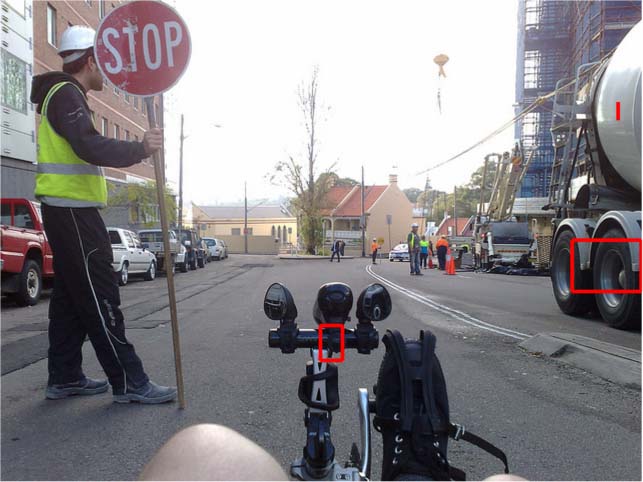}\\
			
			\includegraphics[width=0.10\linewidth]{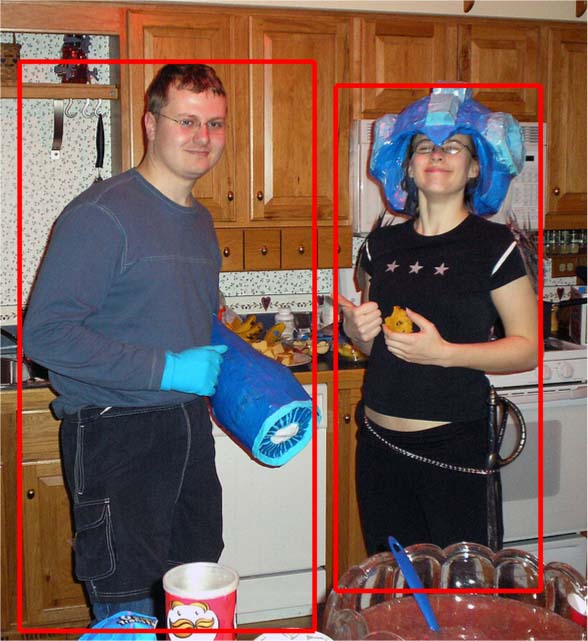}&
			\includegraphics[width=0.10\linewidth]{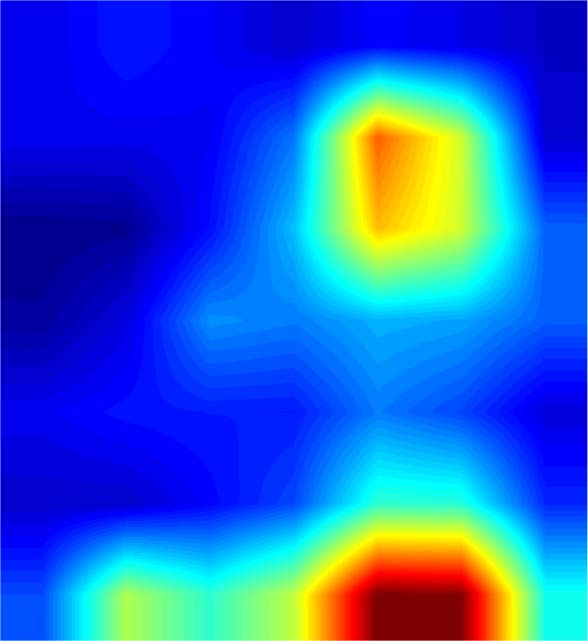}&
			\includegraphics[width=0.10\linewidth]{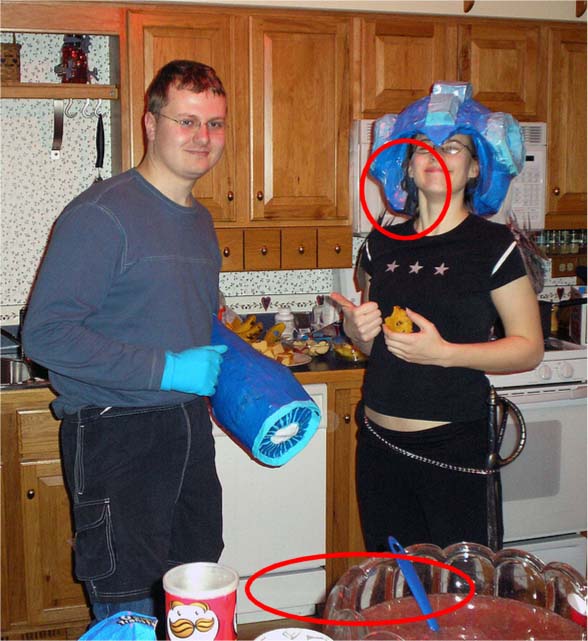}&
			\includegraphics[width=0.10\linewidth]{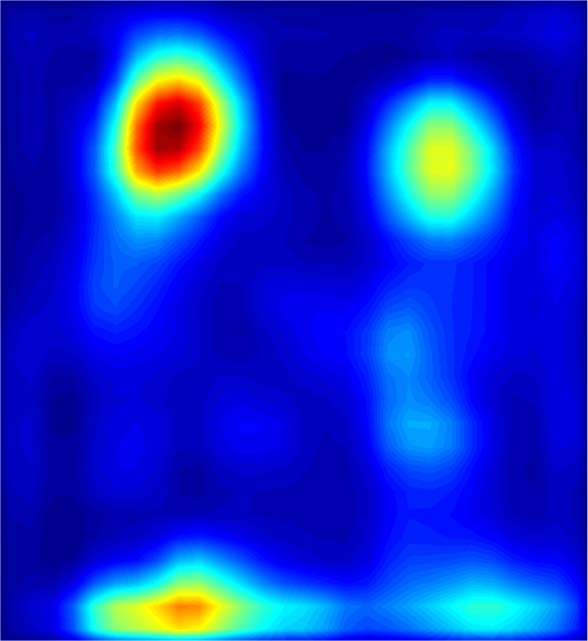}&
			\includegraphics[width=0.10\linewidth]{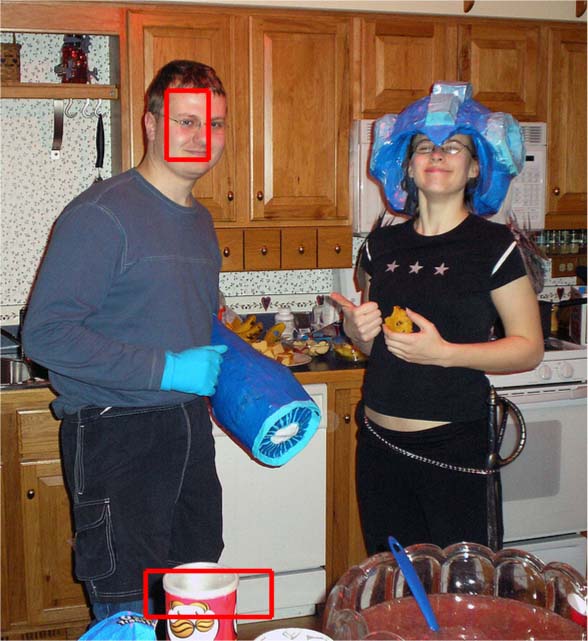}\\
			
			\includegraphics[width=0.10\linewidth]{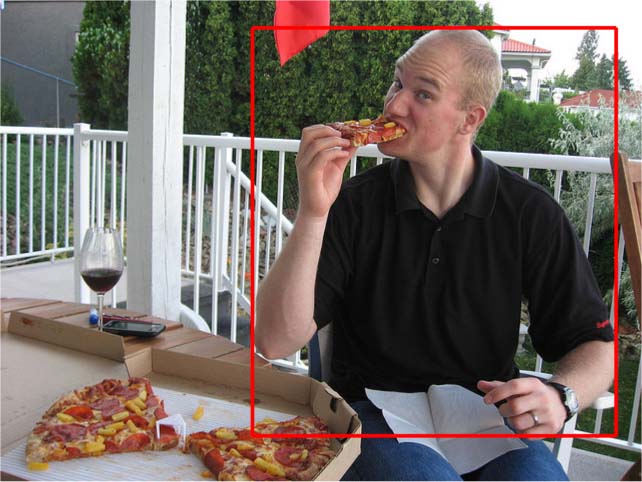}&
			\includegraphics[width=0.10\linewidth]{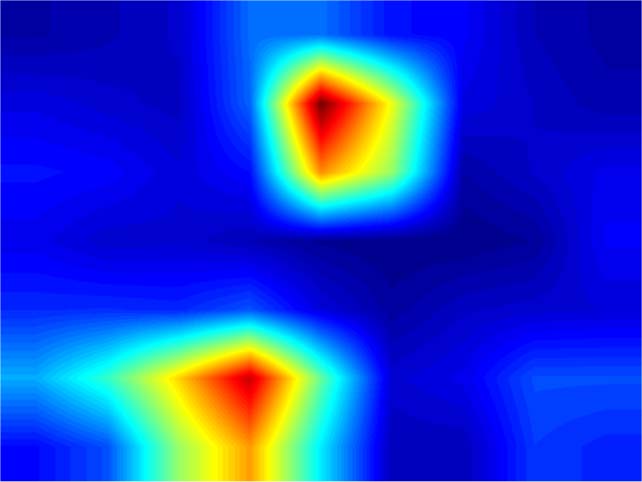}&
			\includegraphics[width=0.10\linewidth]{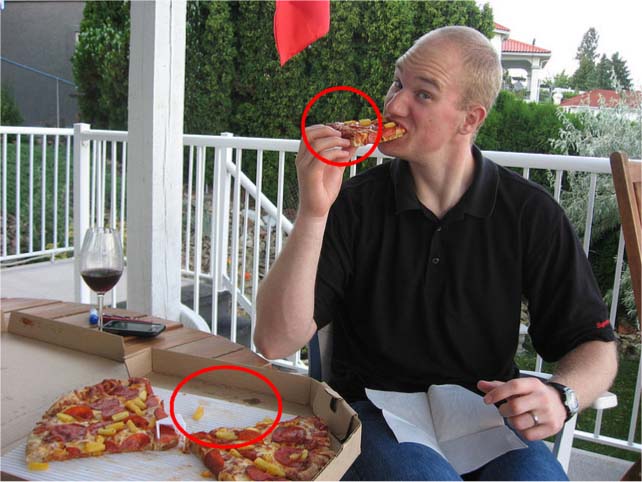}&
			\includegraphics[width=0.10\linewidth]{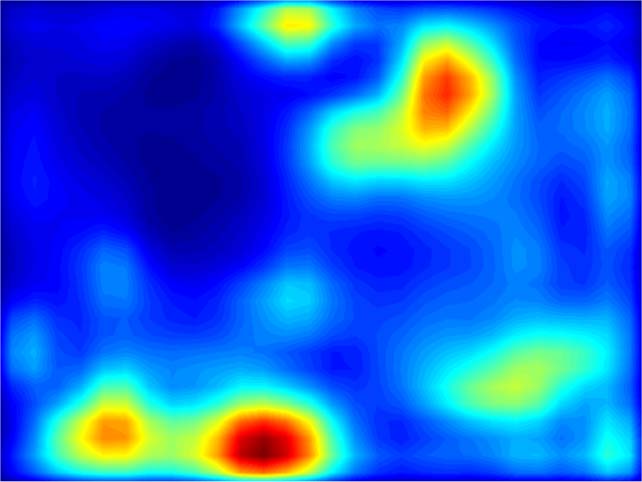}&
			\includegraphics[width=0.10\linewidth]{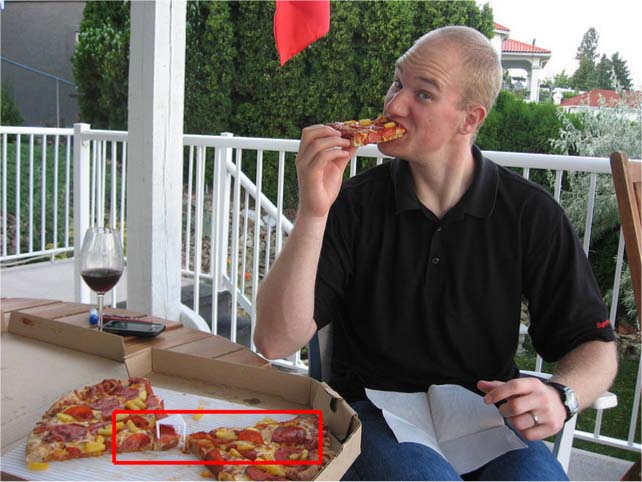}\\
			
			\includegraphics[width=0.10\linewidth]{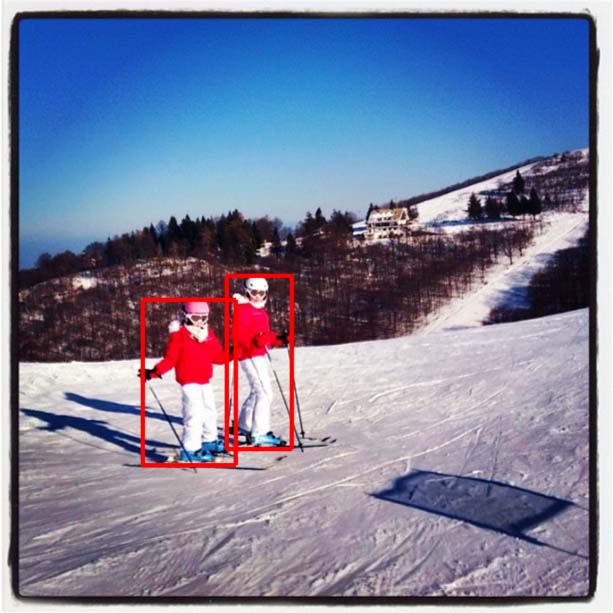}&
			\includegraphics[width=0.10\linewidth]{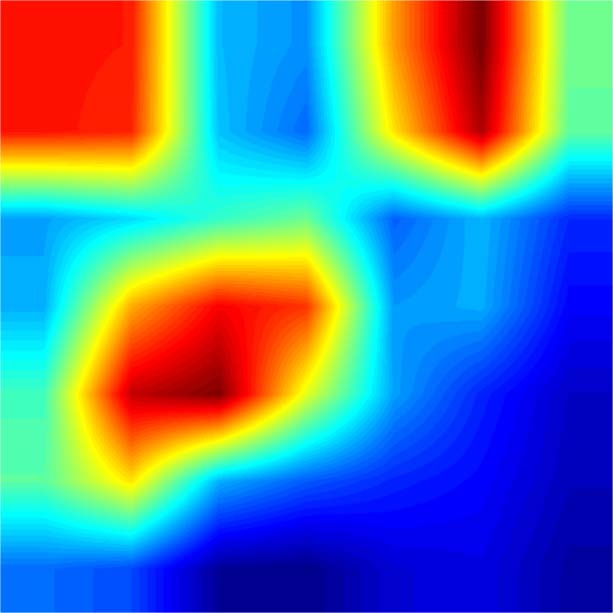}&
			\includegraphics[width=0.10\linewidth]{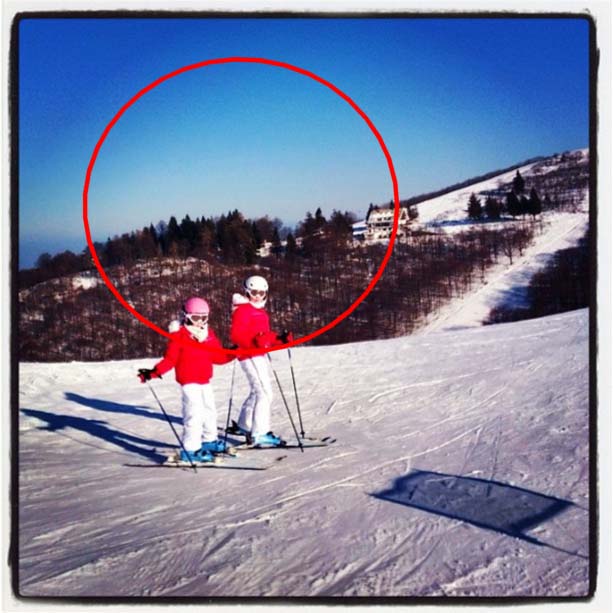}&
			\includegraphics[width=0.10\linewidth]{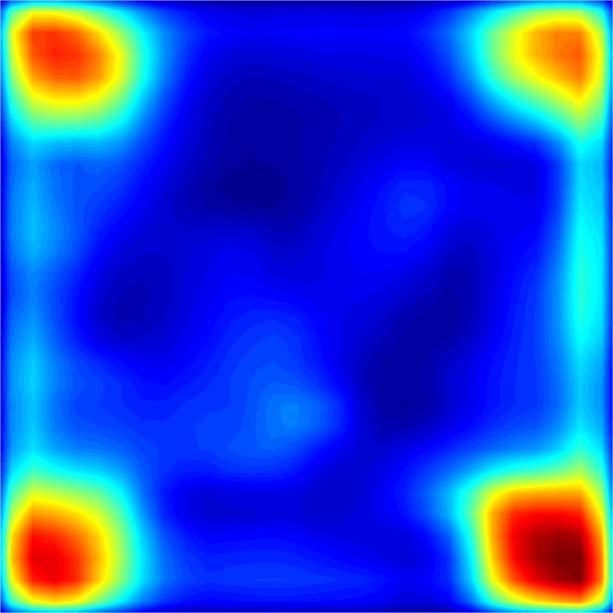}&
			\includegraphics[width=0.10\linewidth]{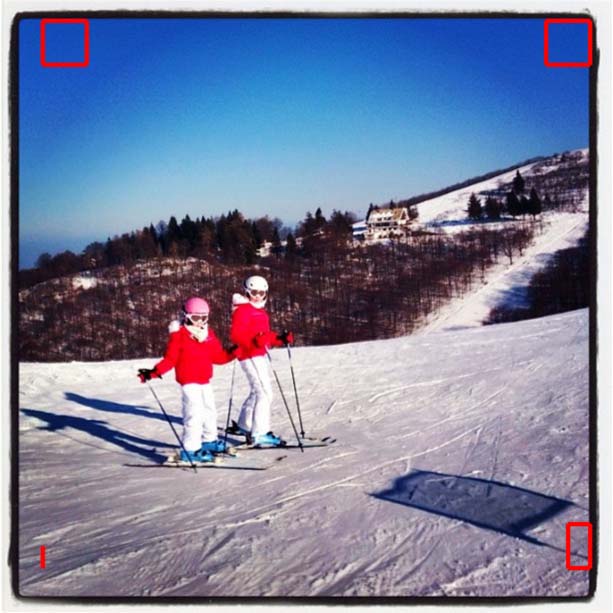}\\
			
			\includegraphics[width=0.10\linewidth]{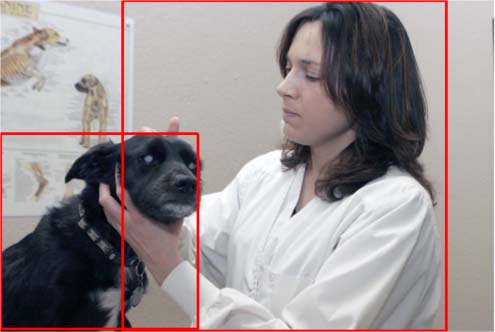}&
			\includegraphics[width=0.10\linewidth]{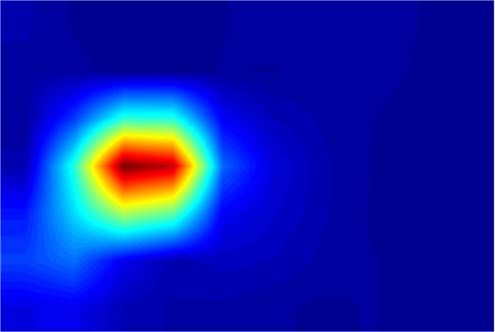}&
			\includegraphics[width=0.10\linewidth]{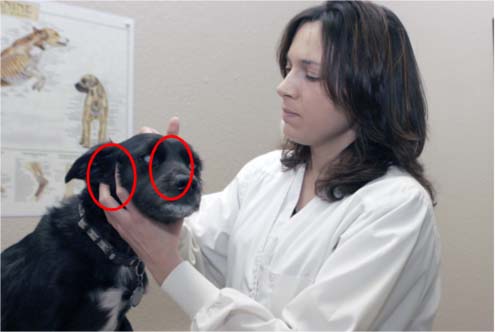}&
			\includegraphics[width=0.10\linewidth]{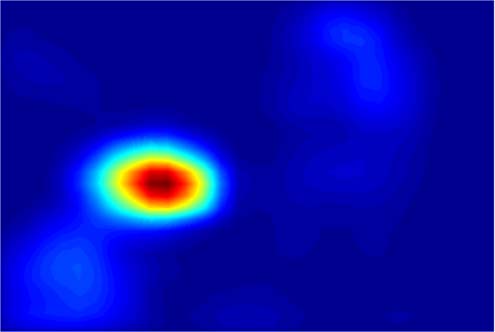}&
			\includegraphics[width=0.10\linewidth]{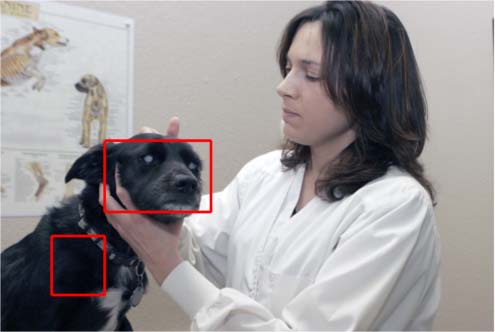}\\
			
			\includegraphics[width=0.10\linewidth]{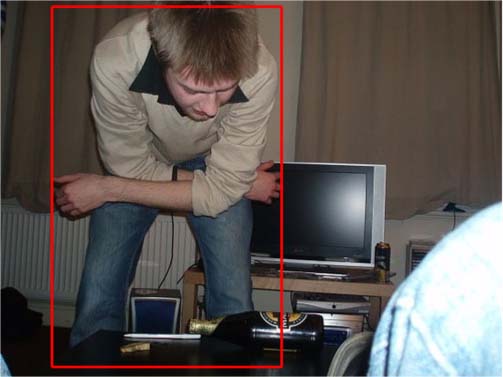}&
			\includegraphics[width=0.10\linewidth]{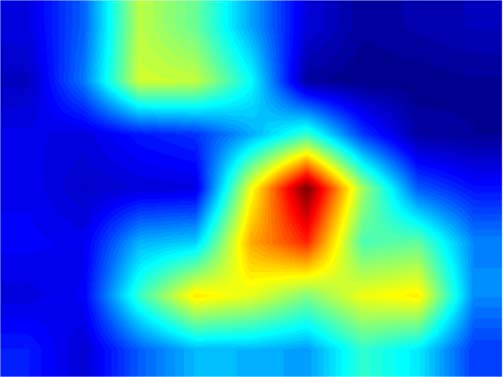}&
			\includegraphics[width=0.10\linewidth]{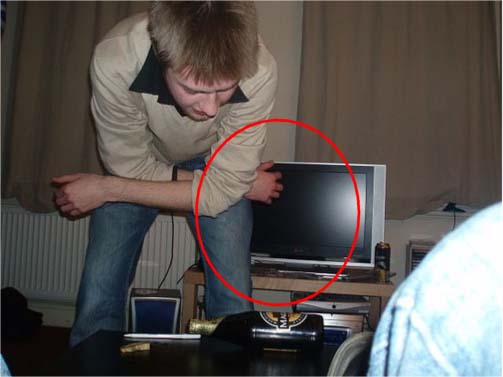}&
			\includegraphics[width=0.10\linewidth]{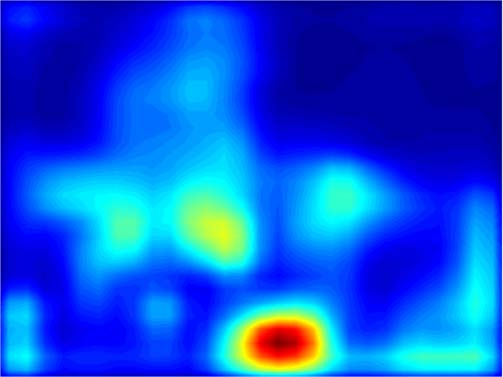}&
			\includegraphics[width=0.10\linewidth]{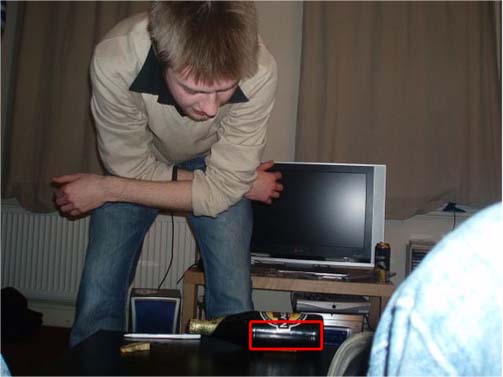}\\
		\end{tabular}
	\end{center}
	\vspace{-3mm}
	\caption{Data bias: the network prefers objects it has classification knowledge}
	\label{tab:Data}
\end{table*}

\begin{table*}[ht]
	\small
	\begin{center}
		\begin{tabular}{ccccc}
			Ground Truth & C2P Heatmap & C2P Output &  P2C Heatmap & P2C Output \\
			\includegraphics[width=0.10\linewidth]{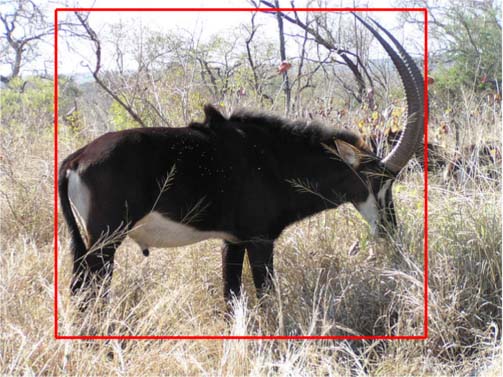}&
			\includegraphics[width=0.10\linewidth]{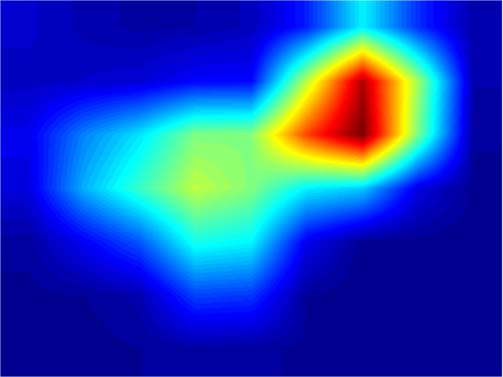}&
			\includegraphics[width=0.10\linewidth]{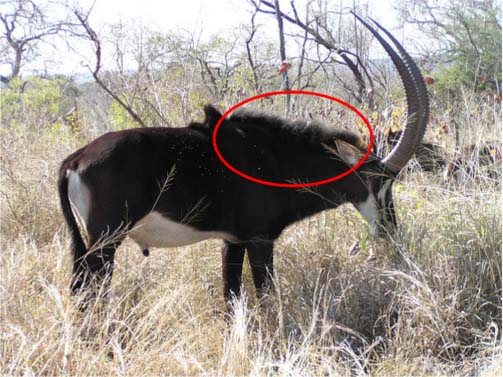}&
			\includegraphics[width=0.10\linewidth]{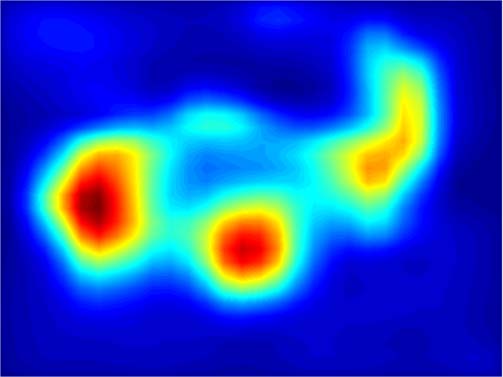}&
			\includegraphics[width=0.10\linewidth]{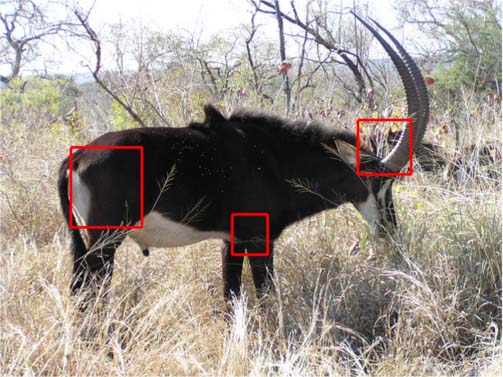}\\
			
			\includegraphics[width=0.10\linewidth]{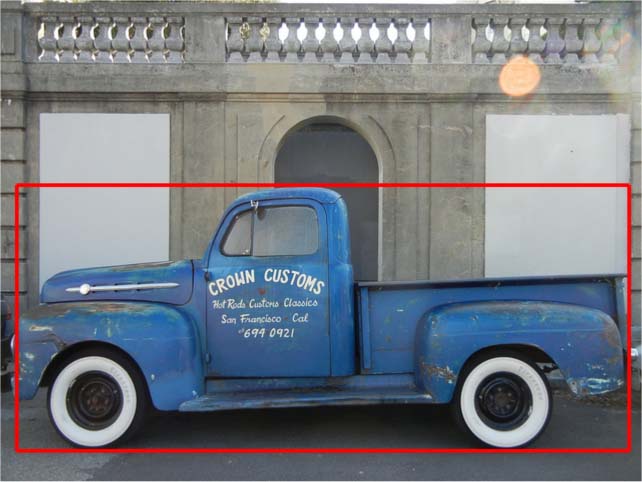}&
			\includegraphics[width=0.10\linewidth]{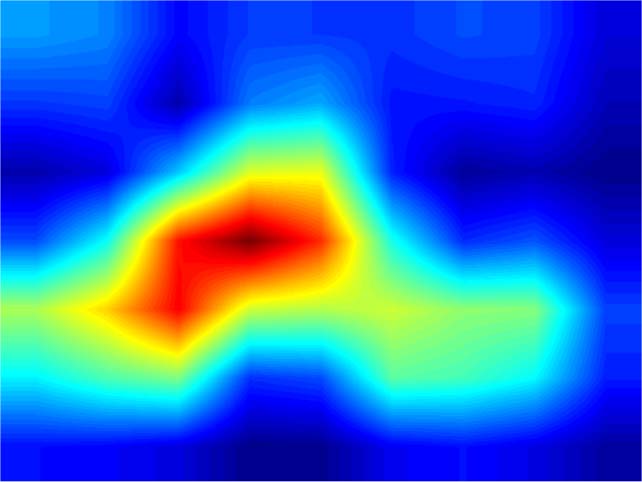}&
			\includegraphics[width=0.10\linewidth]{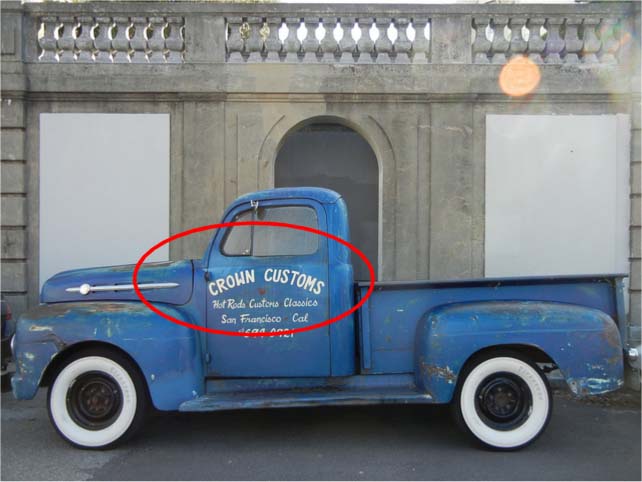}&
			\includegraphics[width=0.10\linewidth]{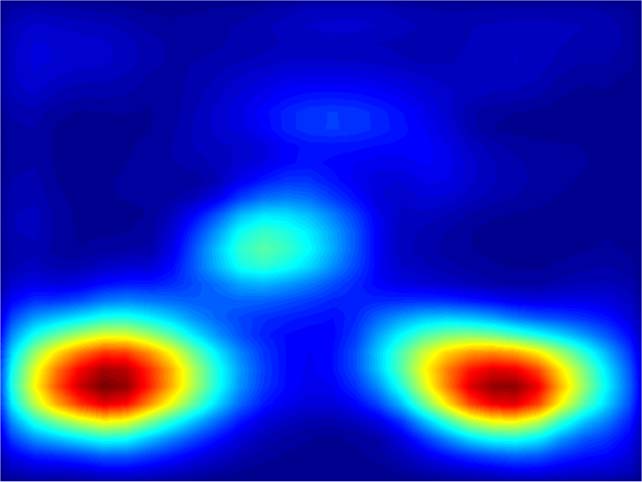}&
			\includegraphics[width=0.10\linewidth]{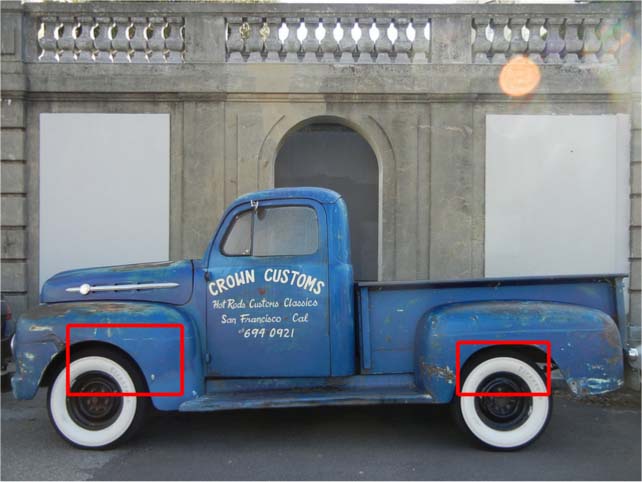}\\
			
			\includegraphics[width=0.10\linewidth]{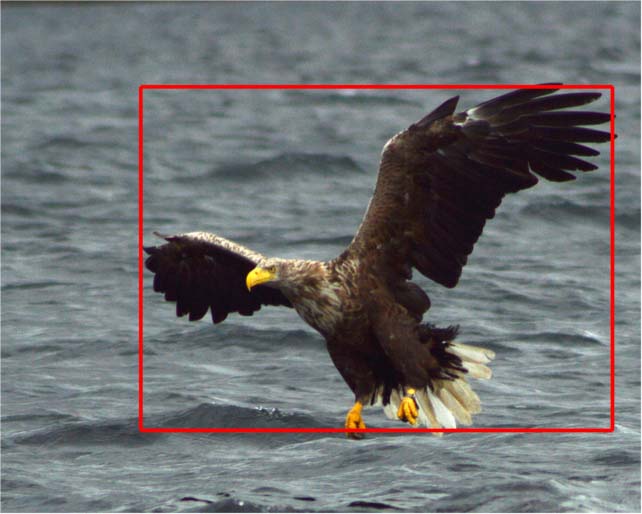}&
			\includegraphics[width=0.10\linewidth]{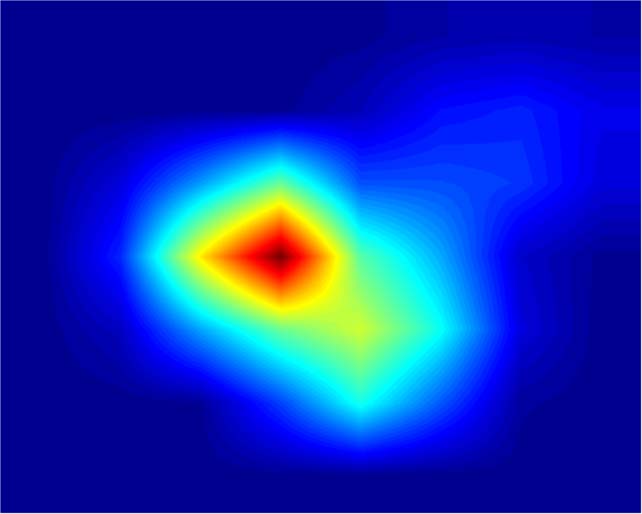}&
			\includegraphics[width=0.10\linewidth]{COCO_COCO_train2014_000000191014_3}&
			\includegraphics[width=0.10\linewidth]{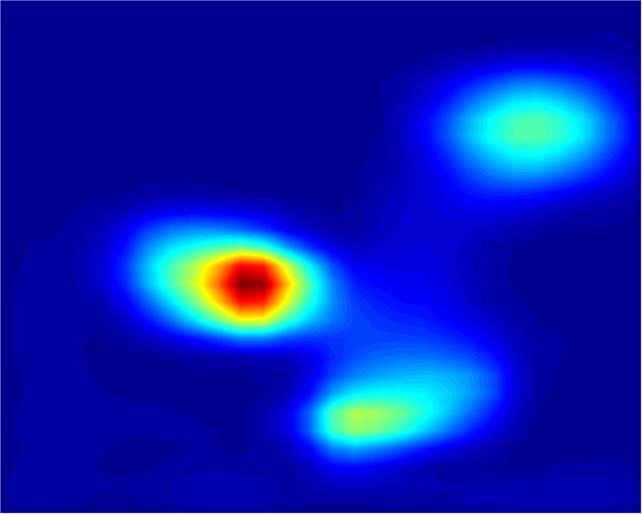}&
			\includegraphics[width=0.10\linewidth]{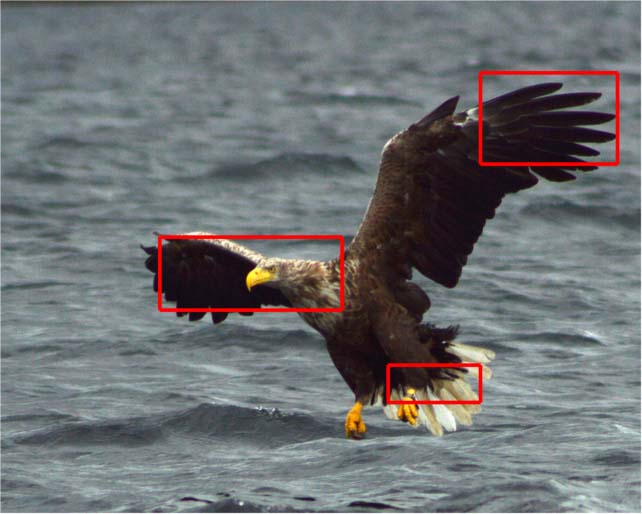}\\
			
			\includegraphics[width=0.10\linewidth]{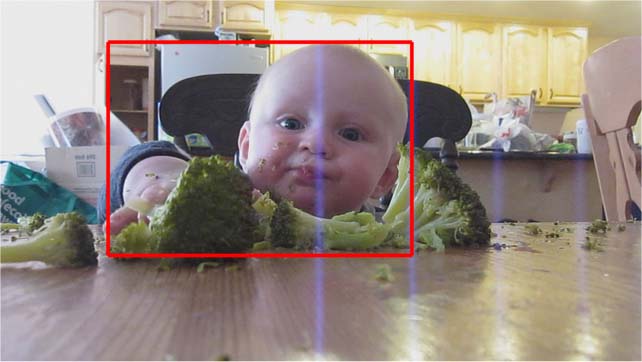}&
			\includegraphics[width=0.10\linewidth]{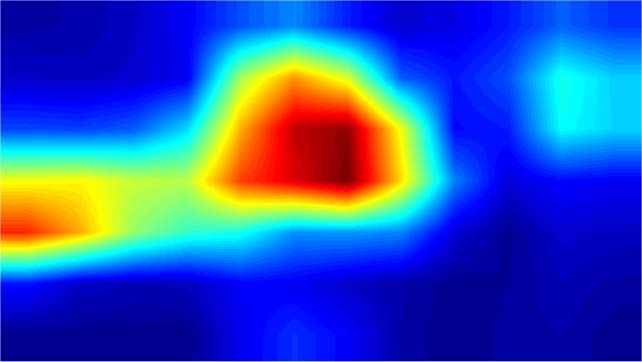}&
			\includegraphics[width=0.10\linewidth]{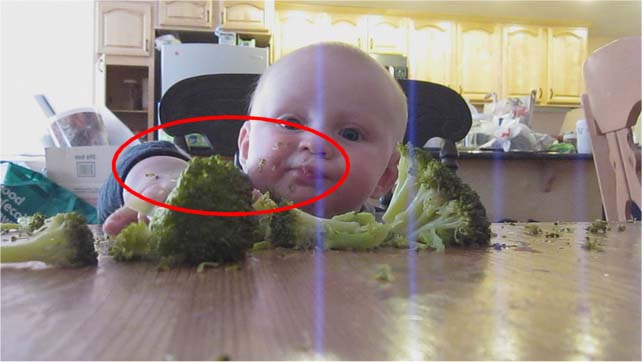}&
			\includegraphics[width=0.10\linewidth]{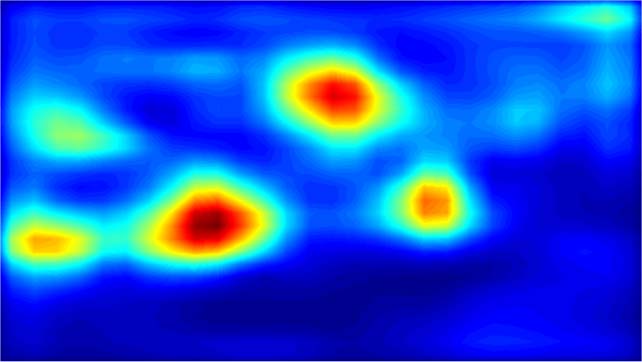}&
			\includegraphics[width=0.10\linewidth]{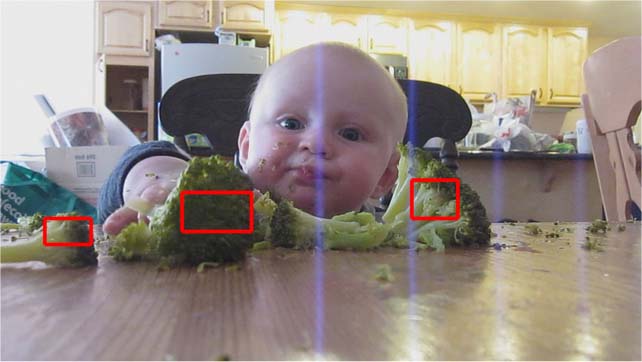}\\

		\end{tabular}
	\end{center}
	\vspace{-3mm}
	\caption{C2P deals with big object better}
	\label{tab:BigO}
\end{table*}

\begin{table*}[ht]
	\small
	\begin{center}
		\begin{tabular}{ccccc}
			Ground Truth & C2P Heatmap & C2P Output &  P2C Heatmap & P2C Output \\
			\includegraphics[width=0.10\linewidth]{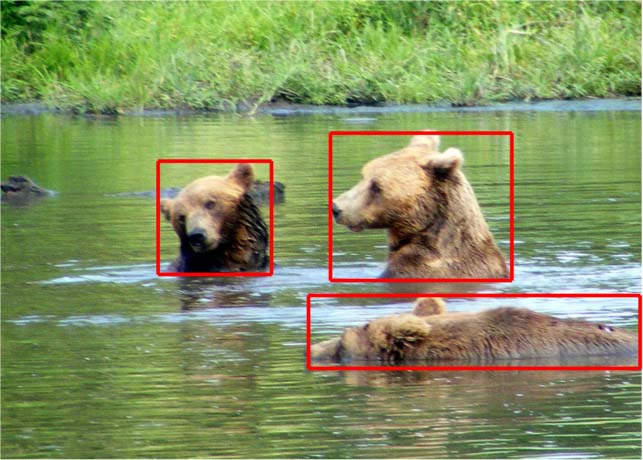}&
			\includegraphics[width=0.10\linewidth]{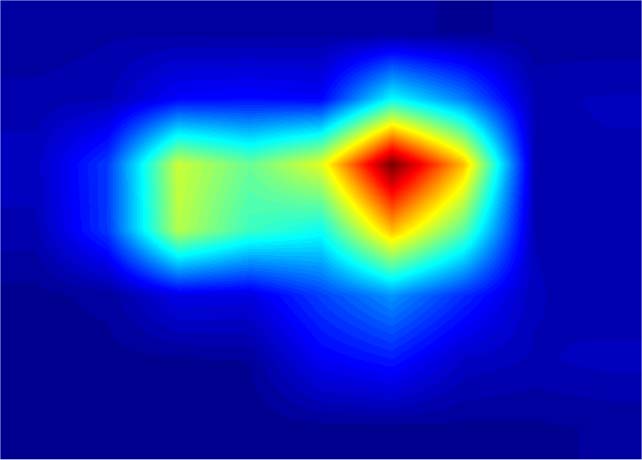}&
			\includegraphics[width=0.10\linewidth]{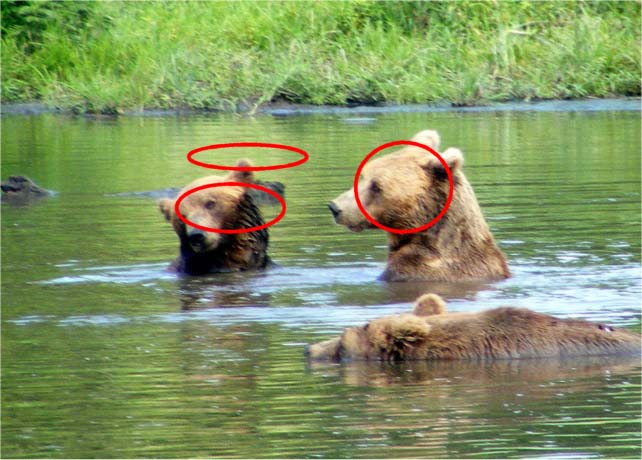}&
			\includegraphics[width=0.10\linewidth]{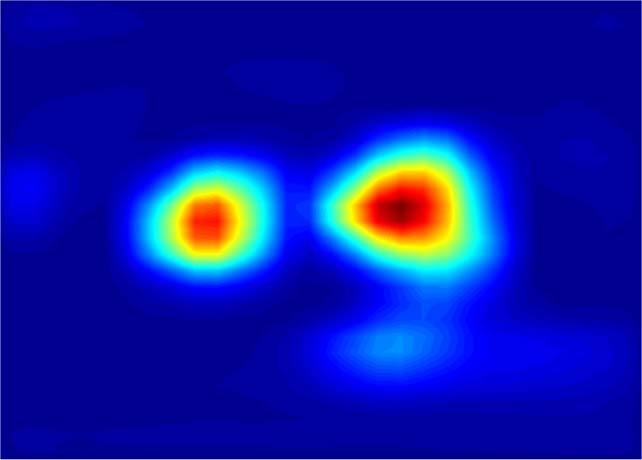}&
			\includegraphics[width=0.10\linewidth]{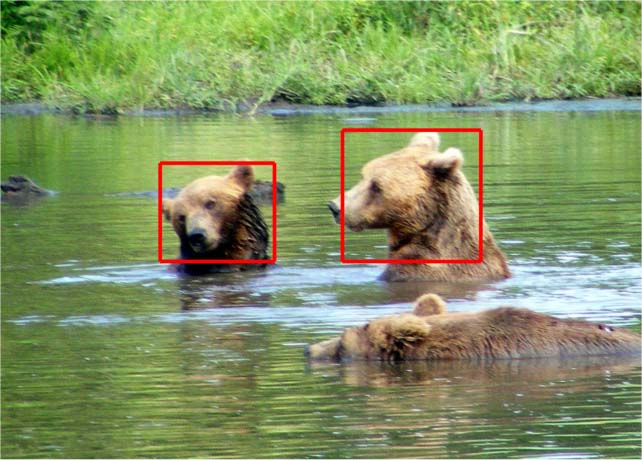}\\
			
			\includegraphics[width=0.10\linewidth]{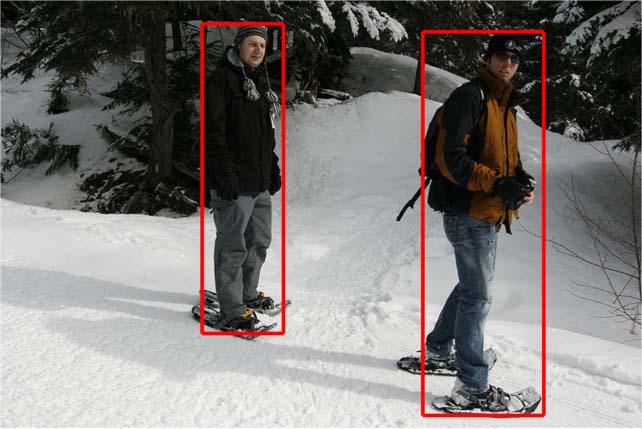}&
			\includegraphics[width=0.10\linewidth]{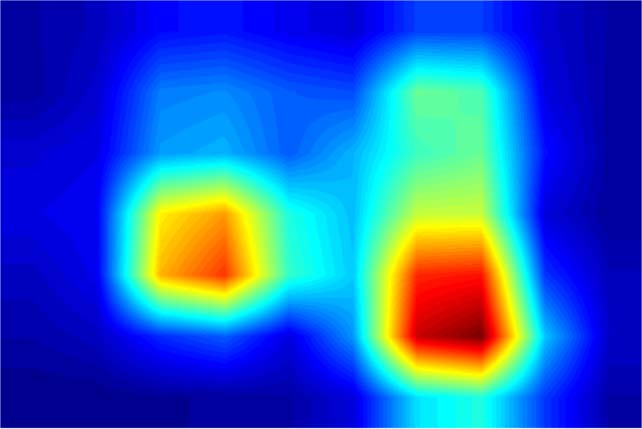}&
			\includegraphics[width=0.10\linewidth]{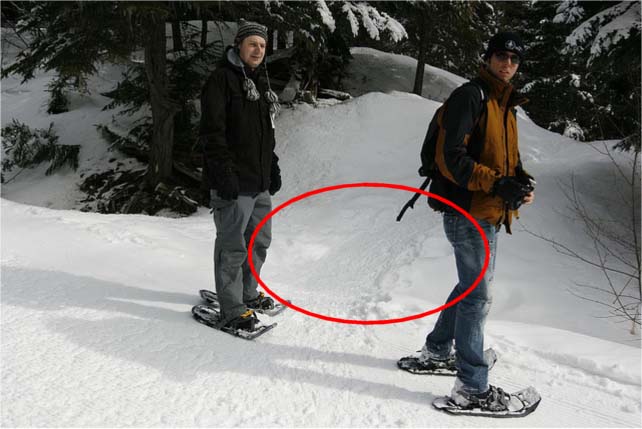}&
			\includegraphics[width=0.10\linewidth]{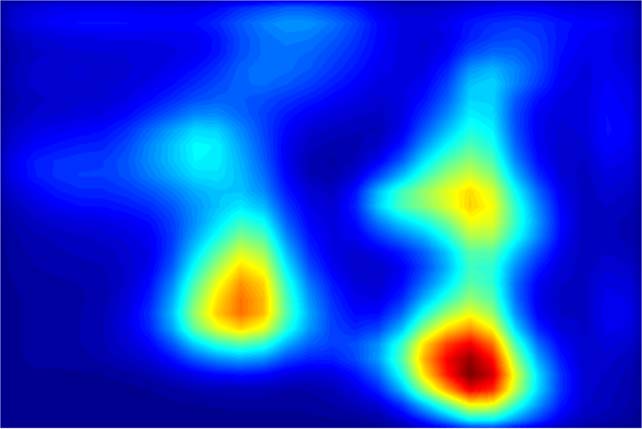}&
			\includegraphics[width=0.10\linewidth]{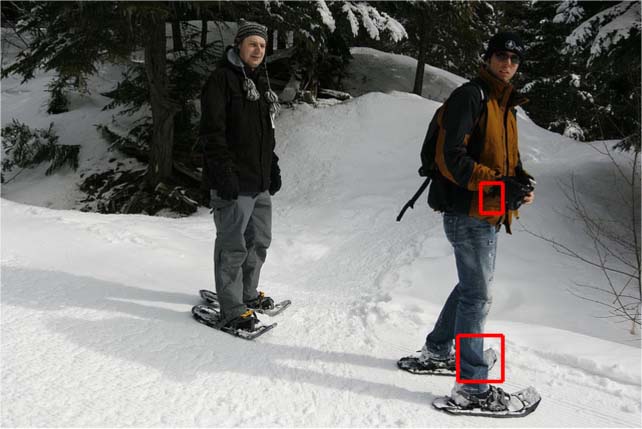}\\
			
			\includegraphics[width=0.10\linewidth]{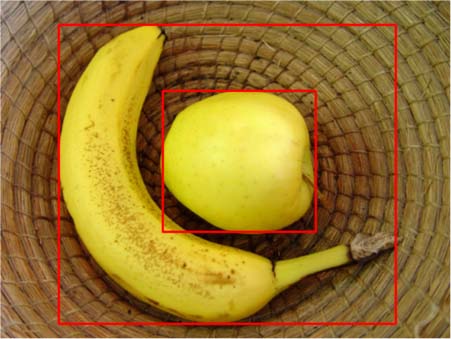}&
			\includegraphics[width=0.10\linewidth]{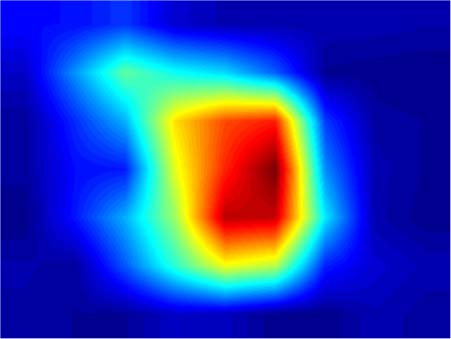}&
			\includegraphics[width=0.10\linewidth]{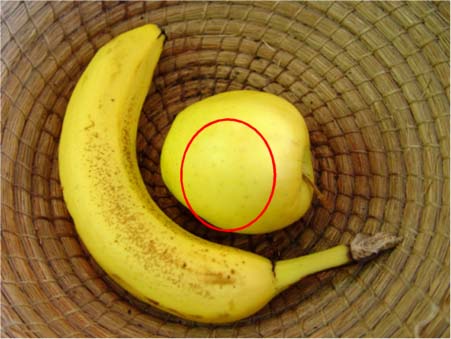}&
			\includegraphics[width=0.10\linewidth]{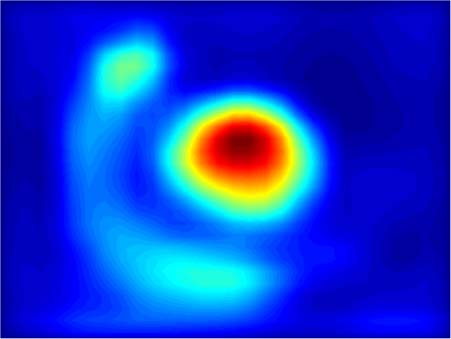}&
			\includegraphics[width=0.10\linewidth]{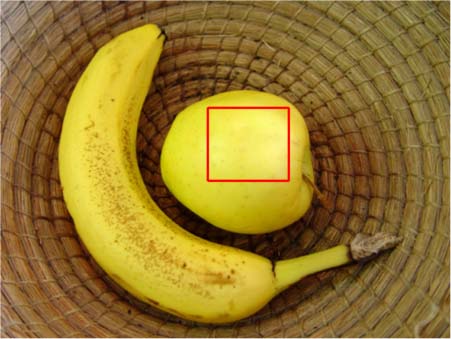}\\
			
			\includegraphics[width=0.10\linewidth]{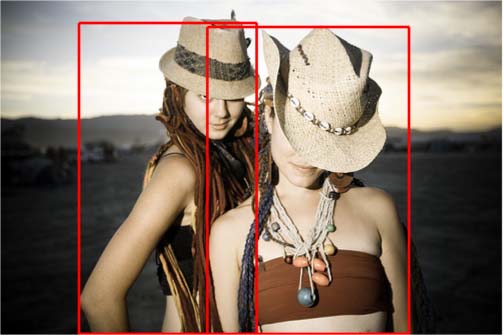}&
			\includegraphics[width=0.10\linewidth]{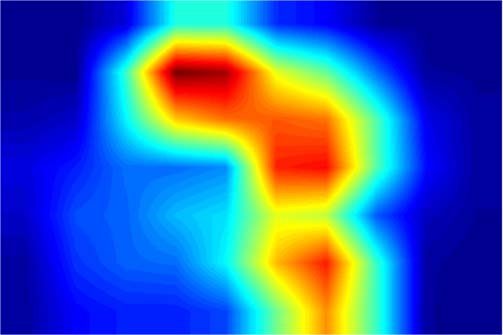}&
			\includegraphics[width=0.10\linewidth]{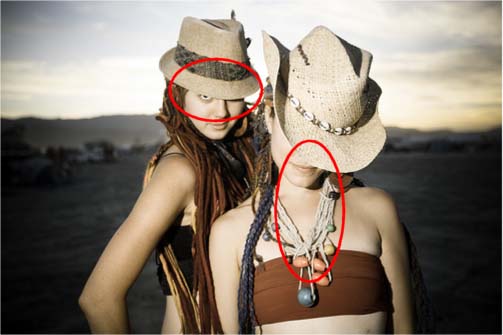}&
			\includegraphics[width=0.10\linewidth]{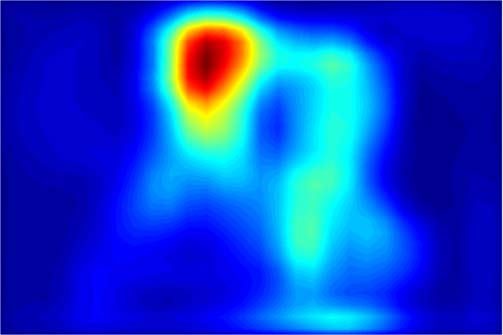}&
			\includegraphics[width=0.10\linewidth]{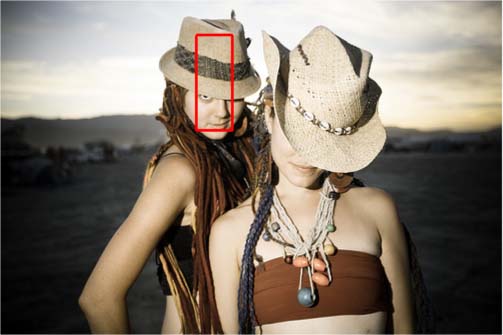}\\

		\end{tabular}
	\end{center}
	\vspace{-3mm}
	\caption{Parallel suppression for mid-sized objects}
	\label{tab:Parallel}
\end{table*}

\begin{table*}[ht]
	\small
	\begin{center}
		\begin{tabular}{ccccc}
			Ground Truth & C2P Heatmap & C2P Output &  P2C Heatmap & P2C Output \\
			\includegraphics[width=0.10\linewidth]{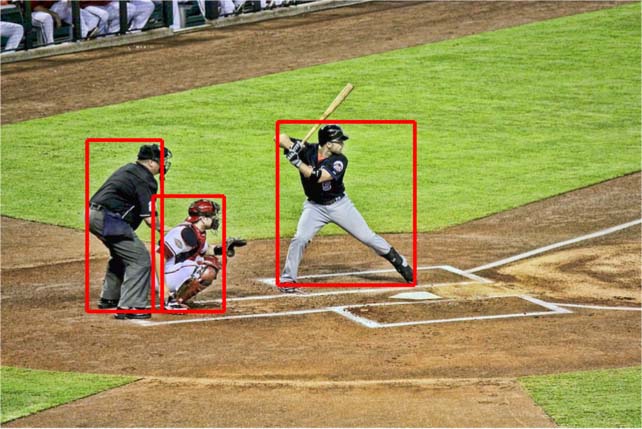}&
			\includegraphics[width=0.10\linewidth]{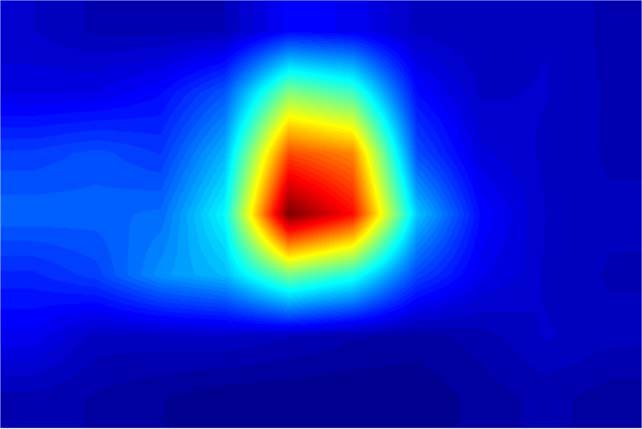}&
			\includegraphics[width=0.10\linewidth]{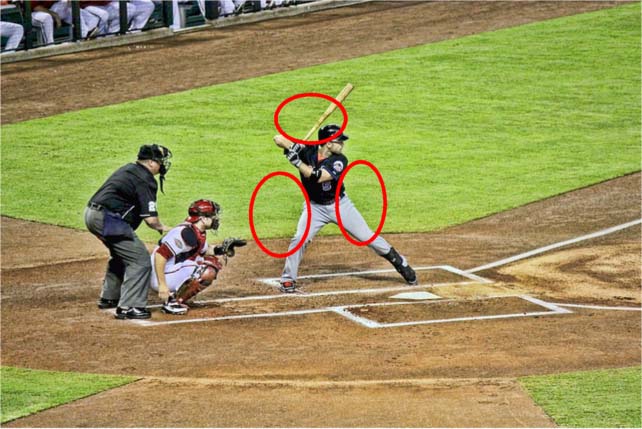}&
			\includegraphics[width=0.10\linewidth]{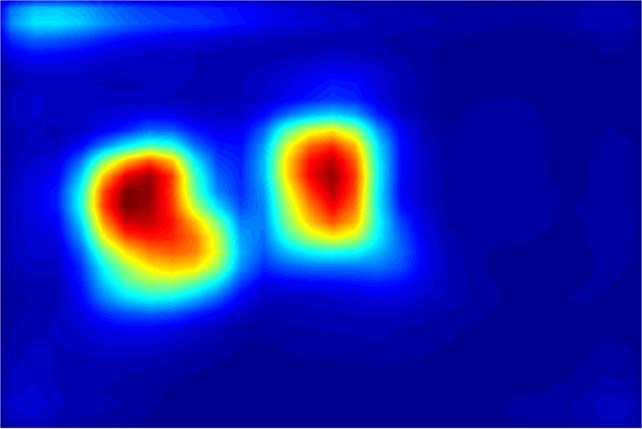}&
			\includegraphics[width=0.10\linewidth]{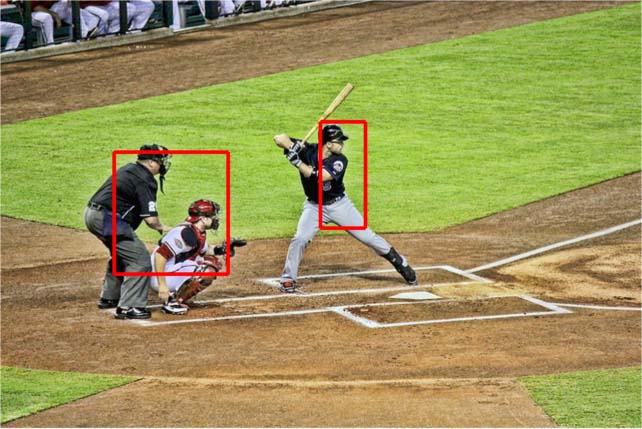}\\
			
			\includegraphics[width=0.10\linewidth]{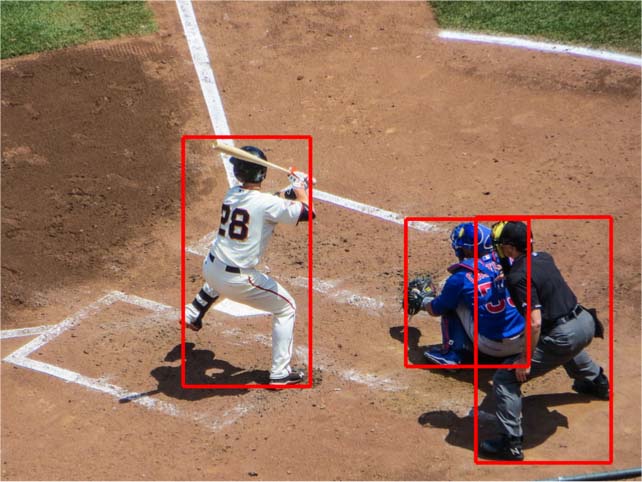}&
			\includegraphics[width=0.10\linewidth]{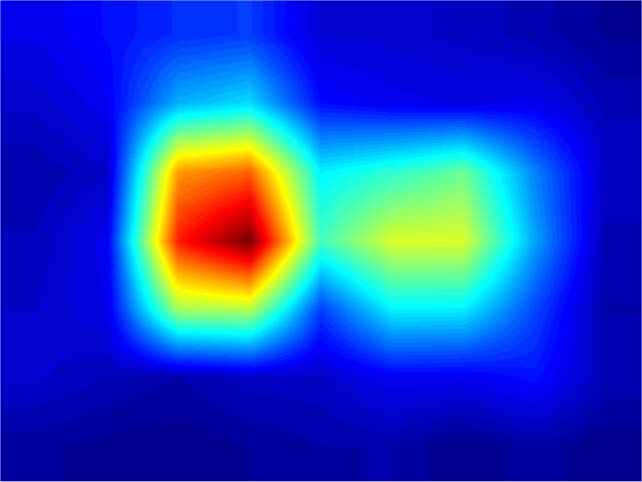}&
			\includegraphics[width=0.10\linewidth]{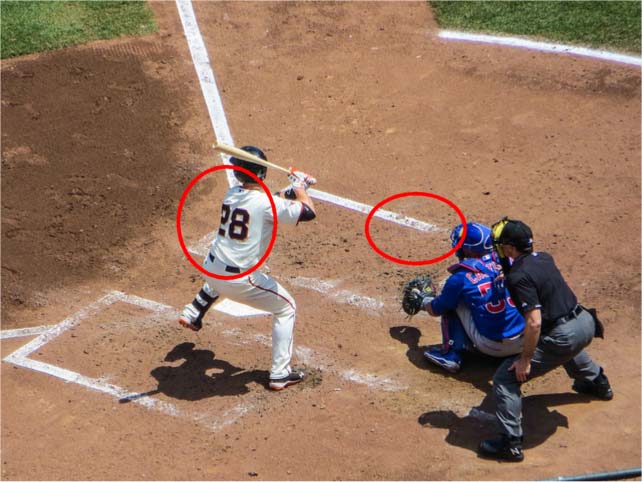}&
			\includegraphics[width=0.10\linewidth]{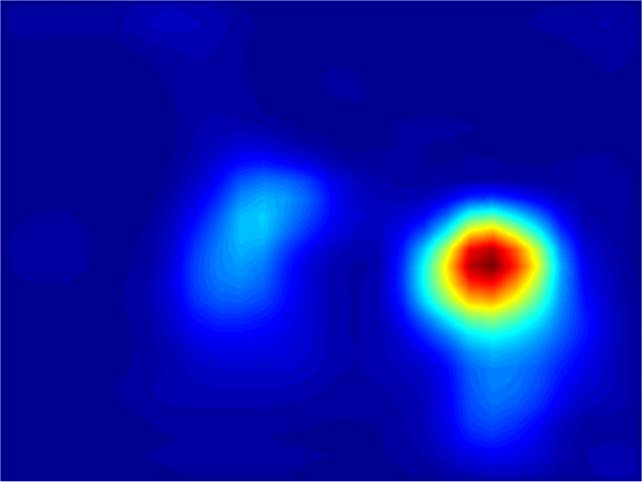}&
			\includegraphics[width=0.10\linewidth]{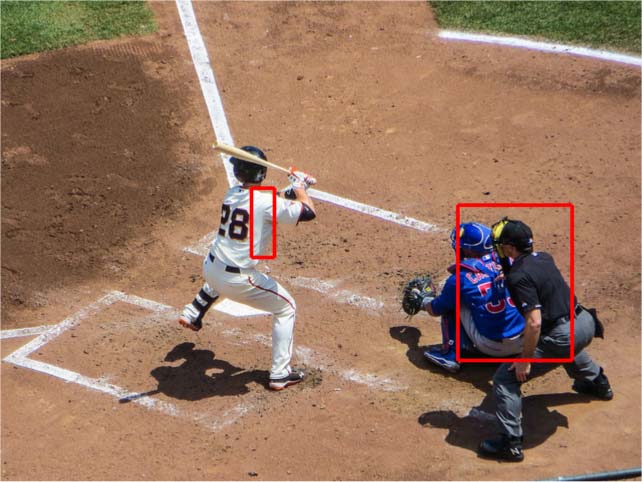}\\
			
			\includegraphics[width=0.10\linewidth]{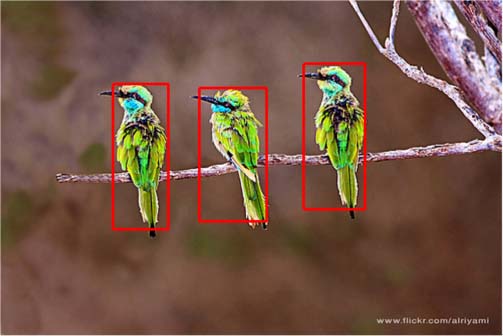}&
			\includegraphics[width=0.10\linewidth]{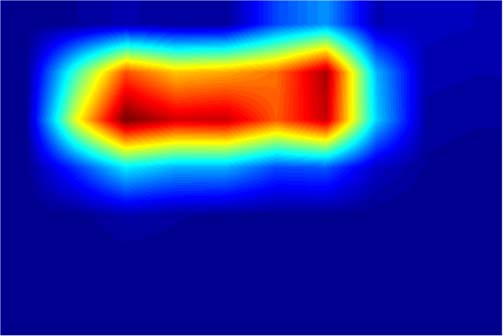}&
			\includegraphics[width=0.10\linewidth]{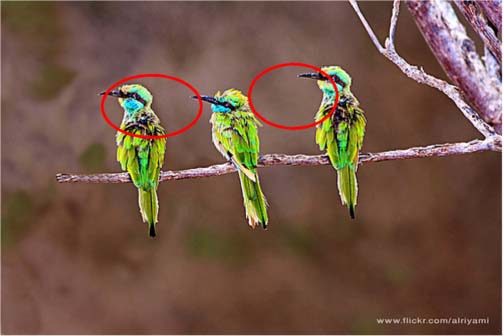}&
			\includegraphics[width=0.10\linewidth]{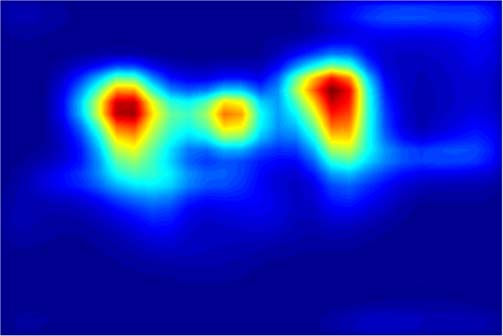}&
			\includegraphics[width=0.10\linewidth]{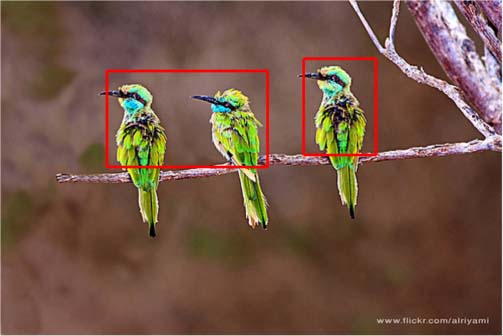}\\
			
			\includegraphics[width=0.10\linewidth]{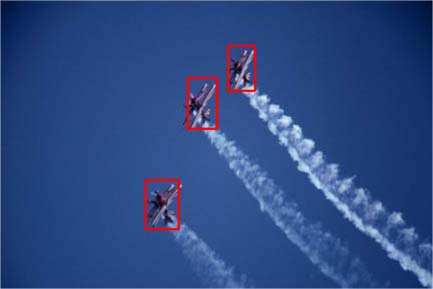}&
			\includegraphics[width=0.10\linewidth]{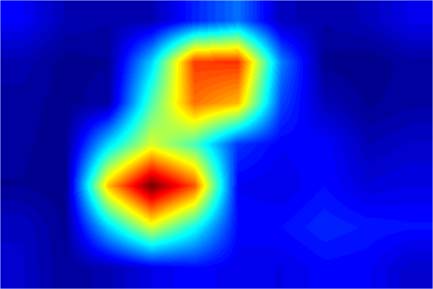}&
			\includegraphics[width=0.10\linewidth]{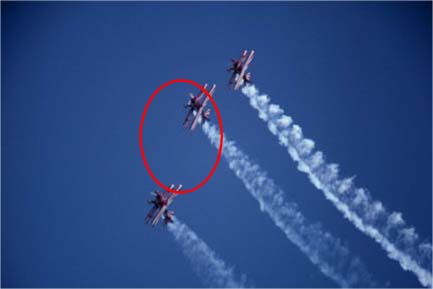}&
			\includegraphics[width=0.10\linewidth]{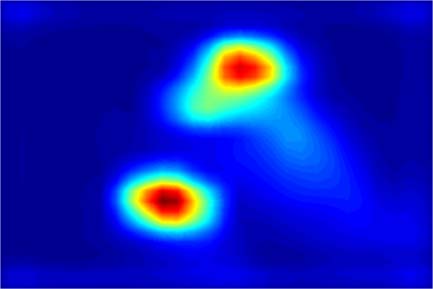}&
			\includegraphics[width=0.10\linewidth]{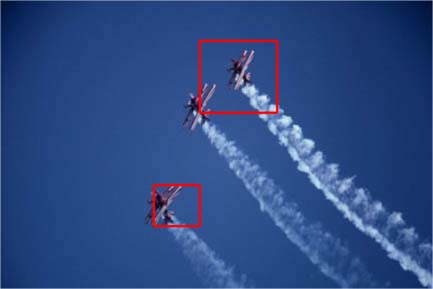}\\
			
		\end{tabular}
	\end{center}
	\vspace{-3mm}
	\caption{Small and close-by objects cause confusion for both approaches}
	\label{tab:Closeby}
\end{table*}
\end{document}